%% file: arxiv.tex
\documentclass[sigconf,nonacm]{acmart}

\usepackage{algorithm}
\usepackage{algpseudocode}
\usepackage{comment}
\usepackage{placeins}
\usepackage{float}
\usepackage{multirow}

\title[MUGEN: Unlearnable Graph Data for Multiple Tasks]{MUGEN: Generating Unlearnable Graph Examples for Multiple Learning Tasks}
\author{Ziyan Liu}
\affiliation{%
  \institution{Harrisburg University of Science and Technology}
  \city{Harrisburg}
  \state{Pennsylvania}
  \country{USA}}
\email{zliu17@my.harrisburgu.edu}
\author{Chengshuai Zhao}
\affiliation{%
  \institution{Arizona State University}
  \city{Tempe}
  \state{Arizona}
  \country{USA}}
\email{czhao93@asu.edu}
\author{Huan Liu}
\affiliation{%
  \institution{Arizona State University}
  \city{Tempe}
  \state{Arizona}
  \country{USA}}
\email{huanliu@asu.edu}

\keywords{unlearnable examples, graph neural networks, multi-task learning}

\newcommand{\posnum}[1]{\underline{#1}}
\newcommand{\negnum}[1]{#1}

\begin{document}

\begin{abstract}
Graph data across diverse domains can expose valuable relational information to unauthorized representation learning, creating a pressing need for protection against such misuse. Unlearnable examples offer a data-level defense by perturbing a training release so that models trained on it fail to generalize to clean data. Existing methods generate unlearnable graph examples for only a specified downstream task. Consequently, a release protected against one task may remain learnable for other plausible uses, including node classification, graph classification, and link prediction, which the data owner cannot anticipate. We introduce MUGEN, to our knowledge the first framework for generating unlearnable graph examples that jointly protect all enabled tasks. From one clean dataset, MUGEN produces a single feature-perturbed release that protects every enabled task through a shared GNN encoder and task-specific heads. We devise a Task-Aligned Separability Objective (TASO), which leverages task prediction and classwise separability to strengthen unlearnability and its transfer across GNN backbones and enabled tasks. We further introduce Type-Adaptive Perturbation (TAP), which tailors perturbation optimization to node-attribute type, with direct search over feasible hard flips that accept only loss-improving updates for discrete node attributes and customized gradient-based updates for continuous node features, thereby enabling strong unlearnability across both settings. Experiments across five benchmarks, four backends and three learning paradigms demonstrate that MUGEN generates transferable unlearnable graph examples across GNN backbones and all three tasks, and remains effective under adversarial training and data augmentation. The code is available at https://github.com/ZiyanLiu16/MUGEN.git.
\end{abstract}

\maketitle

\input{sections/introduction}
\input{sections/related_work}
\input{sections/background_and_problem_formulation}
\input{sections/methodology}
\input{sections/experiments}
\input{sections/conclusion}

\FloatBarrier

\bibliographystyle{ACM-Reference-Format}
\bibliography{references}

\clearpage
\appendix
\input{sections/appendix_1}
\input{sections/appendix}

\end{document}

%% file: sections/introduction.tex
\section{Introduction}

\begin{figure}[t]
    \centering
    \includegraphics[width=0.95\columnwidth]{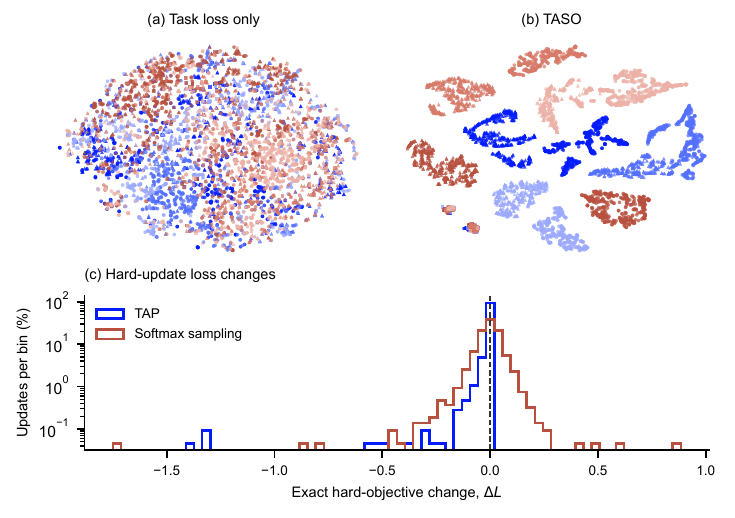}
    \caption{Perturbation representations produced by (a) the task loss and (b) TASO. Color denotes graph class and shape denotes node class. (c) Softmax sampling frequently increases the perturbation loss ($\Delta L>0$), whereas TAP's hard-coordinate updates ensure non-increasing loss ($\Delta L\leq0$).}
    \Description{Three panels compare perturbation representations and categorical-update behavior. The first two show points separated by graph-class color and node-class marker shape; the TASO representation is more class-separated. The third contrasts softmax-sampled updates, some of which increase loss, with hard-coordinate updates whose loss changes are non-positive.}
    \label{fig:introduction-motivation}
\end{figure}

Graphs encode relationships that are central to molecular discovery, scientific networks, social systems, and information retrieval~\citep{sypetkowski2024scalability,wang2024gft,dehghan2024ewek}. Graph neural networks (GNNs) have consequently become a standard learning paradigm for relational data, supporting node classification, graph classification, and link prediction~\citep{kipf2017semi,velickovic2018graph,xu2019powerful,hamilton2017inductive,wang2025tasktrees}. In many settings, however, graph data is owned by one party and released to another. Once released, its attributes, topology, and annotations can be used to train models for purposes not authorized by the data owner. This creates a data-level protection problem: can the owner release graph data whose utility for unauthorized learning is substantially reduced under a specified perturbation budget, while retaining graph topology and task annotations?

Unlearnable examples address this problem by optimizing bounded input perturbations that cause a learner trained on the released data to fail on clean test data~\citep{huang2021unlearnable}. Prior work has strengthened this effect against practical training procedures and studied transfer across architectures and datasets~\citep{fu2022robust,liu2024stable,ren2022transferable}. Graph-data release, however, creates a coverage problem that single-task protection does not resolve: the same graph collection may support node classification, graph classification, or link prediction, and the data owner may not know which task an unauthorized party will pursue. This calls for a one-time protection mechanism whose single released dataset remains unlearnable across its plausible graph-learning uses. However, existing methods for generating unlearnable graph examples~\citep{liu2023unlearnablegraph,liu2023graphcloak} focus on a single downstream task, graph classification, and do not construct one released dataset that jointly protects node-, graph-, and link-level learning.

In addition to this limited task coverage, we identify two empirical challenges that arise when constructing unlearnable examples for graph data using existing methods~\cite{liu2023graphcloak,liu2023unlearnablegraph}. First, optimizing the task loss under the error-minimizing (min--min) poisoning objective adopted by prior graph unlearnability methods produces perturbation representations with weak class separation and does not consistently induce substantial performance degradation across datasets and model backends (Fig.~\ref{fig:introduction-motivation}a). Second, perturbations applied to nodes with categorical attributes generally exhibit substantially weaker unlearnability than those applied to continuous attributes. A significant contributing factor is the inefficient conversion of softmax-sampled updates into discrete perturbations, as adopted in prior methods, since a large fraction of discrete flips do not improve the perturbation objective. For example, on MUTAG, $49.4\%$ of softmax-sampled batch updates increase the perturbation objective (Fig.~\ref{fig:introduction-motivation}c).

To overcome these limitations, we introduce \emph{MUGEN}, a unified framework for generating unlearnable graph data across multiple downstream tasks. Given a clean training dataset, MUGEN produces a single perturbed release that protects every enabled task. It perturbs node features using a shared GNN encoder with task-specific heads and alternates model optimization with joint perturbation updates that combine the enabled task losses with a classwise-separability objective.

Our contributions are fourfold:
\begin{itemize}
    \item We are the first to enable unified protection of node classification, graph classification, and link prediction with a single released dataset, extending beyond prior work focused solely on graph classification.
    \item We devise the \emph{Task-Aligned Separability Objective} (TASO), which introduces classwise-separability signals into perturbation optimization, produces more class-separated perturbation representations (Fig.~\ref{fig:introduction-motivation}b), and strengthens unlearnability and transfer across GNN backbones.
    \item We introduce \emph{Type-Adaptive Perturbation} (TAP), which tailors the optimization steps to the type of node attributes. For categorical attributes, it searches directly over feasible hard flips and accepts only loss-improving updates (Fig.~\ref{fig:introduction-motivation}c), instead of relying on softmax relaxation and sampling, thereby substantially strengthening the unlearnability of the resulting graph examples.
    \item Across five benchmarks covering inductive and transductive settings with continuous and categorical node attributes, we demonstrate that MUGEN produces transferable unlearnable examples. The perturbations transfer across GCN, GAT, GIN, and GraphSAGE, as well as from supervised training to GraphMAE masked autoencoding and GRACE contrastive learning, and remain effective under random feature masking, stochastic edge dropping, and moderate-budget adversarial training.
\end{itemize}

%% file: sections/related_work.tex
\section{Related Work}

\subsection{Unlearnable Example Generation}

Unlearnable examples protect data releases from unauthorized training by inducing models to learn perturbation--label shortcuts rather than task-relevant semantics. \citet{huang2021unlearnable} introduced error-minimizing noise under a bounded perturbation budget; subsequent work improved resistance to adversarial training and stochastic noise~\citep{fu2022robust,liu2024stable} and provided certified measures of reduced learnability~\citep{wang2025provably}. Transferable Unlearnable Examples (TUE) introduced a Classwise Separability Discriminant objective to improve transfer across training settings and datasets~\citep{ren2022transferable}. MTL-UE extends unlearnable-example generation to multi-task image data through a generator conditioned on multiple labels or dense targets associated with each image~\citep{yu2025mtlue}. Generator-based methods broaden the transfer scope beyond fixed datasets and label spaces: One-for-All uses shared image--text concept embeddings for cross-dataset and label-agnostic protection~\citep{chen2024oneforall}, while VTG combines adversarial domain augmentation with text-guided perturbation--label alignment to transfer across domains, label spaces, resolutions, and architectures~\citep{li2025versatile}. Both rely on textual concept or class descriptions during optimization, and their transfer evaluations remain primarily within vision settings. Complementary work improves robustness to practical image-training pipelines: ARMOR targets data augmentation, BAIT counters semantic priors inherited from pretrained backbones, and FUSE distributes protective signals across frequency bands to withstand spectral filtering~\citep{gong2026armor,li2026priors,cai2026fuse}. Methods for generating unlearnable examples have also been adapted to discrete text~\citep{li2023maketext}, time series~\citep{jiang2024timeseries}, 3D point clouds~\citep{wang2024pointclouds}, alignment-triggering protection for LLM fine-tuning~\citep{zhang2026rendering}, and vision--language data for LVLM fine-tuning~\citep{zhao2026mmguard}. Other work broadens the release setting and attacker model: Learnability Lock enables authorized recipients to recover a protected visual release with a secret key~\citep{peng2022learnabilitylock}; CUDA uses class-wise convolutions to generate model-free visual protections that remain effective under several training procedures~\citep{sadasivan2023cuda}; and AUE/AAP target both supervised and contrastive learners~\citep{wang2024efficient}. These methods still operate on visual data and do not address a single graph release that may be repurposed for distinct graph-learning tasks. Unlearnable examples for graphs have been studied through constrained structural perturbations for graph classification~\citep{liu2023unlearnablegraph} and through structural, discrete-feature, and class-wise subgraph perturbations under varying attacker knowledge~\citep{liu2023graphcloak}. However, both methods target graph-level classification and do not produce a single protected release that jointly covers node, graph, and link prediction.

\subsection{Graph Adversarial Attacks}

Graph adversarial attacks share with unlearnable-example generation a common perturbation-based formulation: both modify graph structure or attributes to influence the behavior of graph-learning models. They can be distinguished by attack stage and prediction scope. Evasion attacks modify graph inputs after training to induce incorrect predictions from a fixed model~\citep{dai2018adversarial}. Targeted attacks seek to alter the prediction of a selected node or graph and can be conducted at either inference or training time~\citep{zugner2018adversarial,dai2018adversarial}. Untargeted poisoning instead modifies training data to degrade aggregate performance without requiring the poisoned data to remain easy to fit~\citep{zugner2019adversarial}; related representation-poisoning attacks damage node embeddings used for classification and link prediction and can transfer across learning backends~\citep{bojchevski2019adversarial}. Graph backdoor attacks preserve normal behavior on clean inputs but enforce attacker-chosen predictions when structural or feature triggers are present~\citep{xi2021graphbackdoor,dai2023unnoticeable}. Unlearnable-example generation differs from these attacks in its objectives, mechanisms, and intervention stage and serves as a proactive defense against unauthorized data exploitation~\citep{li2025surveyunlearnable}. Relative to evasion attacks, unlearnable examples are designed at data-release time to prevent models from acquiring task-relevant representations. Compared with untargeted poisoning, they deliberately induce shortcuts that make the protected data appear learnable while impairing generalization to clean data. Unlike backdoor attacks, unlearnable examples adopt an owner-side preventive strategy: protective perturbations are applied broadly to degrade performance across the overall test distribution rather than to control predictions on a trigger-defined subset.

%% file: sections/background_and_problem_formulation.tex
\section{Background and Problem Formulation}
\label{sec:background-problem-formulation}

\subsection{Graph Neural Networks}
\label{subsec:graph-neural-networks}

Let a graph be denoted by \(\mathcal{G}=(\mathcal{V},\mathcal{E},\mathbf{X})\), where \(\mathcal{V}=\{v_1,\ldots,v_N\}\) is the set of \(N\) vertices, \(\mathcal{E}\subseteq\mathcal{V}\times\mathcal{V}\) is the edge set, and \(\mathbf{X}\in\mathbb{R}^{N\times d}\) is the node-feature matrix. Graph connectivity is represented by an adjacency matrix \(\mathbf{A}\in\{0,1\}^{N\times N}\), where \(A_{ij}=1\) if \((v_i,v_j)\in\mathcal{E}\) and \(A_{ij}=0\) otherwise. We write \(\mathcal{N}(v_i)=\{v_j\in\mathcal{V}:A_{ij}=1\}\) for the neighborhood of \(v_i\).

A GNN encoder \(f_\theta\) maps \(\mathcal{G}\) to node representations:
\[
\mathbf{H}=[\mathbf{h}_1,\ldots,\mathbf{h}_N]^{\top}=f_\theta(\mathcal{G}).
\]
The three downstream tasks considered in this work share this encoder but use distinct prediction units:
\begin{equation}
\begin{aligned}
\widehat{\mathbf{y}}_i^{\mathrm{node}}
&=g_{\mathrm{node}}(\mathbf{h}_i), \\
\widehat{\mathbf{y}}^{\mathrm{graph}}
&=g_{\mathrm{graph}}\!\left(\operatorname{READOUT}(\mathbf{H})\right), \\
\widehat{y}_{ij}^{\mathrm{link}}
&=g_{\mathrm{link}}\!\left(\psi(\mathbf{h}_i,\mathbf{h}_j)\right).
\end{aligned}
\label{eq:task-prediction-units}
\end{equation}
Here, ${READOUT}$ denotes the pooling function used to obtain a graph-level representation from the node representations 
and \(\psi\) combines the representations of two candidate link endpoints.  
The three units respectively predict a node label, a graph label, and whether a candidate node pair forms a link. The encoder may instead be pretrained with a self-supervised objective, such as masked-feature reconstruction in GraphMAE~\citep{hou2022graphmae} or contrastive alignment across augmented graph views in GRACE~\citep{zhu2020deep}; the resulting representations can subsequently support the same task-specific prediction units.

\subsection{Problem Formulation}
\label{subsec:problem-formulation}

Let \(\mathcal{D}_{\mathrm{tr}}=\{(\mathcal{G}_k,\mathbf{Y}_k)\}_{k=1}^{n_{\mathrm{tr}}}\) denote clean graph data owned by a data owner, where \(\mathbf{Y}_k\) contains annotations for node, graph, or candidate-pair prediction. The data may comprise a collection of graphs in an inductive setting or one shared graph with training annotations in a transductive setting.

The owner releases a single protected version of the training data, \(\widetilde{\mathcal{D}}_{\mathrm{tr}}=\left\{(\widetilde{\mathcal{G}}_k,\mathbf{Y}_k)\right\}_{k=1}^{n_{\mathrm{tr}}}\), where \(\widetilde{\mathcal{G}}_k=\delta(\mathcal{G}_k)\). Here, \(\delta\) perturbs graph inputs while retaining the annotations required for downstream learning. An unauthorized learner, without access to the clean features or \(\delta\), trains an encoder and task-specific heads on \(\widetilde{\mathcal{D}}_{\mathrm{tr}}\) by optimizing \(\Theta_{\delta}\in\arg\min_{\Theta}\,\mathcal{L}_{\mathrm{train}}\!\left(\Theta;\widetilde{\mathcal{D}}_{\mathrm{tr}}\right)\), where the training loss \(\mathcal{L}_{\mathrm{train}}\) may be supervised or self-supervised.

For intended tasks \(\mathcal{T}\subseteq\{\mathrm{node},\mathrm{graph},\mathrm{link}\}\), our goal is to find a single transformation \(\delta\) such that, for every \(t\in\mathcal{T}\), models trained on \(\widetilde{\mathcal{D}}_{\mathrm{tr}}\) exhibit substantially lower clean-test performance than comparable models trained on \(\mathcal{D}_{\mathrm{tr}}\).

%% file: sections/methodology.tex
\section{Methodology}\label{sec:methodology}

\subsection{Supervised Multi-Task Unlearnable Graph Generation}
\label{subsec:multi-task-generation}

\begin{figure*}[t]
    \centering
    \includegraphics[width=\textwidth]{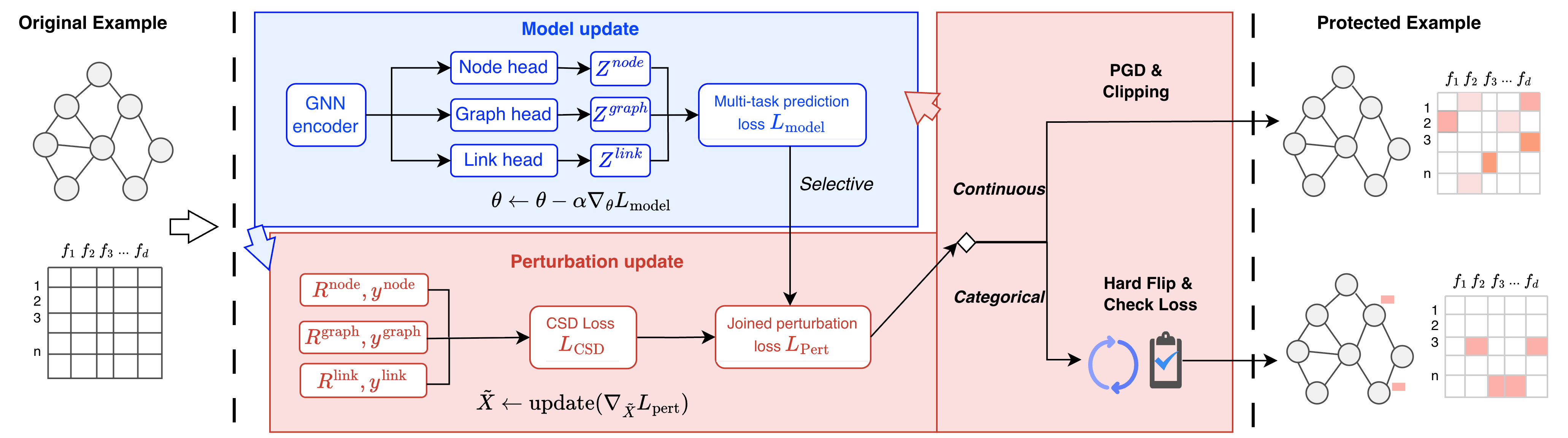}
    \caption{Overview of MUGEN. We alternate GNN updates using the enabled-task prediction losses and perturbation updates. Continuous attributes use clipped PGD, whereas categorical attributes use hard flips with exact loss evaluation.}
    \Description{A workflow diagram shows a clean graph dataset feeding a shared GNN encoder and task-specific heads for graph, node, and link prediction. Prediction losses update the model. A joint perturbation objective then updates continuous features by clipped projected gradient descent or categorical features by hard-coordinate flips with exact evaluation, producing the released perturbed dataset.}
    \label{fig:framework-overview}
\end{figure*}

We instantiate the transformation \(\delta\) from the problem formulation as a feature-only perturbation. For each released graph, \(\widetilde{\mathcal{G}}_k=(\mathcal{V}_k,\mathcal{E}_k,\widetilde{\mathbf{X}}_k)\): the node-feature matrix \(\widetilde{\mathbf{X}}_k\) is perturbed, whereas the \(\mathcal{V}_k\), \(\mathcal{E}_k\), and \(\mathcal{Y}_k\) are retained. We train a GNN encoder with task-specific heads for the enabled tasks in \(\mathcal{T}\) on the perturbed features, and alternately update the model parameters and feature perturbations. Thus, every perturbation update is shaped jointly by the prediction units that the released graph data is intended to protect. Section~\ref{subsec:perturbation-parameterization} defines \(\widetilde{\mathbf{X}}\), Section~\ref{subsec:multi-task-objective} defines the losses, and Section~\ref{subsec:alternating-optimization} specifies the alternating updates.
The overall workflow is summarized in Figure~\ref{fig:framework-overview}.

\subsection{Node-Feature Perturbation Parameterization}
\label{subsec:perturbation-parameterization}

Let \(\mathbf{M}\in\{0,1\}^{N\times 1}\) select nodes whose attributes may be changed. In inductive settings, these are nodes in training graphs; in transductive settings, they are the training-mask nodes.

\paragraph{Continuous attributes.}
For continuous node attributes, we optimize an additive perturbation \(\boldsymbol{\Delta}\in\mathbb{R}^{N\times d}\):
\begin{equation}
\widetilde{\mathbf{X}}
=\mathbf{X}+\mathbf{M}\odot\boldsymbol{\Delta},
\qquad
\Delta_{ij}\in[-\varepsilon_j,\varepsilon_j].
\label{eq:continuous-perturbation}
\end{equation}
The feature-wise radius is \(\varepsilon_j=\rho\sigma_j\), where \(\rho\) is the perturbation-budget ratio and \(\sigma_j\) is the standard deviation of feature dimension \(j\) over the training data.

\paragraph{Categorical attributes.}
For categorical node attributes, \(\mathbf{x}_i\) is a one-hot vector and \(\widetilde{\mathbf{x}}_i\) must remain one-hot. We therefore perturb attributes through hard category reassignment. Following GraphCloak's node-attribute perturbation budget~\citep{liu2023graphcloak}, for a graph \(g\) with \(V_g\) nodes and \(E_g\) non-self edges, we impose
\begin{equation}
\sum_{i\in\mathcal{V}_g}
\mathbb{I}\!\left[\widetilde{\mathbf{x}}_i\ne\mathbf{x}_i\right]
\leq B_g,
\qquad
B_g=\left\lfloor\min\!\left(\frac{E_g}{10},\,\beta V_g^2\right)\right\rfloor,
\label{eq:categorical-perturbation}
\end{equation}
where \(\beta\) controls the graph-wise flip budget. This preserves valid categorical attributes while limiting the number of altered nodes.

\subsection{Task-Specific Objectives and Multi-Task Perturbation Loss}
\label{subsec:multi-task-objective}

TASO constructs a single perturbation to jointly protect all enabled tasks in
$\mathcal{T}$. Its perturbation objective is centered on task-aligned
classwise separability, using the \emph{Classwise Separability Discriminant}
(CSD), originally introduced to improve the transferability of unlearnable
examples~\citep{ren2022transferable}. The conventional task losses can
additionally be included as auxiliary terms, allowing TASO to retain the
prior task-loss signal while introducing a separability-based objective for
transferable protection.

For each task, TASO applies CSD either to learned representations or to the
feature perturbations. In representation mode, given encoder outputs
$\mathbf{Z}$, we use
\[
\mathbf{R}^{\mathrm{node}}=\mathbf{Z},\qquad
\mathbf{R}^{\mathrm{graph}}_g
=\operatorname{READOUT}(\mathbf{Z}_{\mathcal{V}_g}),\qquad
\mathbf{r}^{\mathrm{link}}_{uv}
=\psi(\mathbf{z}_u,\mathbf{z}_v).
\]
In perturbation mode, the corresponding representations are constructed from
$\widetilde{\mathbf{X}}-\mathbf{X}$. Each $\mathbf{R}^{t}$ is paired with its
task labels $\mathbf{y}^{t}$.

\subsubsection{Task Loss Construction}
\label{subsubsec:task-loss}

For each enabled task $t\in\mathcal{T}$, let
$\mathcal{L}^{\mathrm{task}}_t$ denote its ordinary supervised prediction
loss: cross-entropy for node and graph classification, and binary
cross-entropy for link prediction. Prior graph unlearnable-example methods
optimize this loss for a single downstream task using an error-minimizing
min--min objective~\citep{liu2023unlearnablegraph,liu2023graphcloak}:
\begin{equation}
\delta_t^{\star}
\in
\arg\min_{\delta\in\mathcal{C}}
\min_{\Theta}
\mathcal{L}^{\mathrm{task}}_t
\!\left(
\Theta;\delta(\mathcal{D}_{\mathrm{tr}})
\right).
\label{eq:prior-task-min-min}
\end{equation}
Thus, the resulting perturbation is optimized for one selected downstream
task. In contrast, TASO aggregates their task losses into the weighted objective 
\begin{equation}
\mathcal{L}_{\mathrm{model}}
=
\sum_{t\in\mathcal{T}}
w_t\mathcal{L}^{\mathrm{task}}_t.
\label{eq:model-loss}
\end{equation}
This objective is used for learner-parameter optimization. Its inclusion in
the perturbation objective is optional and is specified separately below.

\subsubsection{Task-Aligned Classwise Separability}
\label{subsubsec:task-aligned-csd}

The perturbation update additionally uses the CSD. For each enabled task, CSD is computed from the
task-specific pair
$(\mathbf{R}^{t},\mathbf{y}^{t})$ defined above. For a generic collection of
vector--label pairs $\{(\mathbf{r}_i,y_i)\}_{i=1}^{N}$, let
\[
\mathcal{I}_c=\{i:y_i=c\},
\qquad
\boldsymbol{\mu}_c
=
\frac{1}{|\mathcal{I}_c|}
\sum_{i\in\mathcal{I}_c}\mathbf{r}_i,
\]
and define the mean intra-class dispersion as
\[
s_c
=
\frac{1}{|\mathcal{I}_c|}
\sum_{i\in\mathcal{I}_c}
\left\lVert\mathbf{r}_i-\boldsymbol{\mu}_c\right\rVert_2.
\]
The corresponding CSD loss is
\begin{equation}
\mathcal{L}_{\mathrm{CSD}}(\mathbf{R},\mathbf{y})
=
\frac{1}{C(C-1)}
\sum_{c\ne c'}
\frac{s_c+s_{c'}+\epsilon}
{\left\lVert\boldsymbol{\mu}_c-\boldsymbol{\mu}_{c'}\right\rVert_2+\epsilon},
\label{eq:csd-loss}
\end{equation}
where $C$ is the number of observed classes and $\epsilon>0$ is a numerical
stability constant. Compared with the prior task-loss-only objective, CSD
explicitly shapes task-relevant classwise structure and is designed to
improve perturbation transferability.

The perturbation objective combines the primary task-aligned CSD terms with
the optional task-loss signal:
\begin{equation}
\mathcal{L}_{\mathrm{pert}}
=
\lambda_{\mathrm{task}}\mathcal{L}_{\mathrm{model}}
+
\lambda_{\mathrm{CSD}}
\sum_{t\in\mathcal{T}}
w_t
\mathcal{L}_{\mathrm{CSD}}
\left(\mathbf{R}^{t},\mathbf{y}^{t}\right),
\label{eq:multi-task-perturbation-loss}
\end{equation}
where $\lambda_{\mathrm{task}}\ge 0$ and
$\lambda_{\mathrm{CSD}}>0$. Setting $\lambda_{\mathrm{task}}=0$ gives the
CSD-only variant, while $\lambda_{\mathrm{task}}>0$ retains the conventional
task-loss signal as an auxiliary perturbation objective.

\subsection{Alternating Optimization}
\label{subsec:alternating-optimization}

At iteration \(r\), we first update the model parameters on the current perturbed input:
\begin{equation}
\boldsymbol{\Theta}^{(r)}
\leftarrow
\boldsymbol{\Theta}^{(r-1)}
-\alpha\nabla_{\boldsymbol{\Theta}}
\mathcal{L}_{\mathrm{model}}
\left(
\boldsymbol{\Theta}^{(r-1)};
\widetilde{\mathbf{X}}^{(r-1)}
\right).
\label{eq:model-update}
\end{equation}
We then freeze \(\boldsymbol{\Theta}^{(r)}\) and update the perturbation within the feasible space defined in Section~\ref{subsec:perturbation-parameterization}. For continuous attributes, TAP uses projected gradient descent (PGD)~\citep{madry2018towards}; for categorical attributes, it performs a hard-coordinate search.

Both perturbation updates optimize
$\mathcal{L}_{\mathrm{pert}}(\boldsymbol{\Theta}^{(r)};\widetilde{\mathbf{X}})$
within the feasible space while keeping
$\boldsymbol{\Theta}^{(r)}$ fixed. The resulting perturbed features are then
used in the next learner update in Equation~\ref{eq:model-update} and, after the final iteration, form the
released protected data. The continuous and categorical update procedures
are described below.

\subsubsection{Continuous attributes: PGD}
We use a sign gradient or row-wise \(\ell_2\)-normalized gradient, followed by projection onto the feature-wise box constraint:
\begin{equation}
\boldsymbol{\Delta}^{(r+1)}
=\operatorname{clip}
\left(
\boldsymbol{\Delta}^{(r)}
-\eta_{\mathrm{pgd}}\operatorname{sign}
\left(\nabla_{\boldsymbol{\Delta}}\mathcal{L}_{\mathrm{pert}}\right),
-\boldsymbol{\varepsilon},
\boldsymbol{\varepsilon}
\right).
\label{eq:continuous-pgd-update}
\end{equation}

\subsubsection{Categorical attributes: hard-coordinate search.}
Prior relaxed categorical optimization~\citep{liu2023graphcloak} maintains logits $\mathbf a_i\in\mathbb R^{C_{\mathrm{cat}}}$ and probabilities $\mathbf p_i=\operatorname{softmax}(\mathbf a_i)$. Deployment samples $S_i\sim\operatorname{Cat}(\mathbf p_i)$ and applies the graph-wise budgets $\{B_g\}$ from Equation~\ref{eq:categorical-perturbation}, yielding hard features $\widehat{\mathbf X}=\Pi_{\mathrm{bud}}(\mathbf S)$, where $\Pi_{\mathrm{bud}}$ denotes budget enforcement. The relaxed update follows a surrogate gradient $\widetilde{\mathbf g}(\mathbf a)$, but in general 
\[
\widetilde{\mathbf{g}}(\mathbf{a})
\not\equiv
\nabla_{\mathbf{a}}\,
\mathbb{E}_{\mathbf{S}\sim\operatorname{Cat}(\mathbf{p})}
\!\left[
\mathcal{L}_{\mathrm{pert}}
\!\left(\boldsymbol{\Theta}^{(r)};\Pi_{\mathrm{bud}}(\mathbf{S})\right)
\right].
\]
Consequently, a descent step under the relaxation need not decrease the objective of the sampled, budget-feasible hard features. TAP instead searches directly in the discrete feasible set. Inspired by HotFlip's~\citep{ebrahimi-etal-2018-hotflip} first-order scoring of atomic one-hot flips, we use the gradient only to rank budget-feasible hard changes and evaluate the resulting candidates with the exact perturbation objective.
With \(\boldsymbol{\Theta}^{(r)}\) fixed, let \(c_i\) be the current category of node \(i\), \([C_{\mathrm{cat}}]\) the category set, and \(\mathbf{g}_i=\nabla_{\widetilde{\mathbf{x}}_i}\mathcal{L}_{\mathrm{pert}}(\boldsymbol{\Theta}^{(r)};\widetilde{\mathbf{X}})\). For a reassignment \(c_i\!\rightarrow\!c\), \(c\ne c_i\), the first-order predicted objective change is
\(s_{i,c}=[\mathbf{g}_i]_c-[\mathbf{g}_i]_{c_i}\).
Thus, lower \(s_{i,c}\) predicts a larger objective decrease. At coordinate step \(q\), Algorithm~\ref{alg:categorical-hard-coordinate} selects at most one budget-feasible operation \(\omega_g^\star\) per graph: a label reassignment or a swap that reverts one existing flip and adds another. The negative-score operations are collected into a multi-operation plan \(\mathcal{P}^{(q)}\). Let \(\mathcal C^{(q)}\) contain the hard input produced by this full plan and those produced by up to \(K_{\mathrm{exact}}\) lowest-score singleton operations. We select
\begin{equation}
\widetilde{\mathbf X}^{(q+1)}
\in
\arg\min_{\mathbf X'\in\{\widetilde{\mathbf X}^{(q)}\}\cup\mathcal C^{(q)}}
\mathcal L_{\mathrm{pert}}\!\left(\boldsymbol{\Theta}^{(r)};\mathbf X'\right),
\label{eq:categorical-exact-acceptance}
\end{equation}
which guarantees
\[
\begin{aligned}
\Delta\mathcal L_{\mathrm{hard}}^{(q)}
={}&\mathcal L_{\mathrm{pert}}\!\left(\boldsymbol{\Theta}^{(r)};\widetilde{\mathbf X}^{(q+1)}\right)\\
&-\mathcal L_{\mathrm{pert}}\!\left(\boldsymbol{\Theta}^{(r)};\widetilde{\mathbf X}^{(q)}\right)\leq0.
\end{aligned}
\]
This gradient-guided comparison avoids enumerating all subsets of \(\mathcal{P}^{(q)}\) and requires at most \(K_{\mathrm{exact}}+1\) exact evaluations per coordinate step.

\begin{algorithm}[!t]
\caption{Categorical Hard-Coordinate Subroutine}
\label{alg:categorical-hard-coordinate}
\begin{algorithmic}[1]
\Require frozen \(\boldsymbol{\Theta}^{(r)}\), current hard features \(\widetilde{\mathbf{X}}^{(0)}\), batch \(\mathcal{B}\), graph budgets \(\{B_g\}_{g\in\mathcal{B}}\), coordinate steps \(Q\), singleton limit \(K_{\mathrm{exact}}\)
\Ensure updated hard features \(\widetilde{\mathbf{X}}^{(Q)}\)
\For{\(q=0,\ldots,Q-1\)}
    \State \(\mathbf{G}\gets\nabla_{\widetilde{\mathbf{X}}}\mathcal{L}_{\mathrm{pert}}(\boldsymbol{\Theta}^{(r)};\widetilde{\mathbf{X}}^{(q)})\)
    \For{each graph \(g\in\mathcal{B}\)}
        \For{each active node \(i\in\mathcal{V}_g\)}
            \For{each \(c\in[C_{\mathrm{cat}}]\setminus\{c_i\}\)}
                \If{\(c_i\!\rightarrow\!c\) is feasible under \(B_g\)}
                    \State \(s_{i,c}\gets[\mathbf{G}_i]_c-[\mathbf{G}_i]_{c_i}\)
                \EndIf
            \EndFor
        \EndFor
        \State \(\omega_g^\star\gets\) lowest-score label reassignment or swap that reverts one existing flip and adds another
    \EndFor
    \State \(\mathcal{P}^{(q)}\gets\{\omega_g^\star:\,g\in\mathcal{B},\ s(\omega_g^\star)<0\}\)
    \State \(\begin{aligned}
        \mathcal{C}^{(q)}\gets{}&\{\operatorname{Apply}(\widetilde{\mathbf{X}}^{(q)},\mathcal{P}^{(q)})\}\\[-1pt]
        &{}\cup\{\operatorname{Apply}(\widetilde{\mathbf{X}}^{(q)},\{\omega\}):\\[-1pt]
        &\qquad\omega\in\operatorname{TopK}(\mathcal{P}^{(q)},K_{\mathrm{exact}})\}
    \end{aligned}\)
    \State \(\widehat{\mathbf{X}}\gets\arg\min_{\mathbf{X}\in\mathcal{C}^{(q)}}\mathcal{L}_{\mathrm{pert}}(\boldsymbol{\Theta}^{(r)};\mathbf{X})\)
    \If{\(\mathcal{L}_{\mathrm{pert}}(\boldsymbol{\Theta}^{(r)};\widehat{\mathbf{X}})<\mathcal{L}_{\mathrm{pert}}(\boldsymbol{\Theta}^{(r)};\widetilde{\mathbf{X}}^{(q)})\)}
        \State \(\widetilde{\mathbf{X}}^{(q+1)}\gets\widehat{\mathbf{X}}\)
    \Else
        \State \(\widetilde{\mathbf{X}}^{(q+1)}\gets\widetilde{\mathbf{X}}^{(q)}\)
    \EndIf
\EndFor
\State \Return \(\widetilde{\mathbf{X}}^{(Q)}\)
\end{algorithmic}
\end{algorithm}

Together, the learner update in Equation~\ref{eq:model-update} and the
type-specific perturbation updates in Equations~\ref{eq:continuous-pgd-update}
and~\ref{eq:categorical-exact-acceptance} define the alternating optimization
procedure used to generate the final feasible perturbed release.

%% file: sections/experiments.tex
\section{Experimental Evaluation}

\subsection{Experimental Settings}

\begin{table}[t]
\centering
\caption{Scale and learning regimes of the datasets used in our evaluation.}
\label{tab:dataset-task-coverage}
\begin{tabular}{lcc}
\toprule
Dataset & Avg. nodes & Regime \\
\midrule
MUTAG & 17.9 & Inductive \\
ENZYMES & 32.6 & Inductive \\
PROTEINS\_full & 39.1 & Inductive \\
Cora & 2,708 & Transductive \\
PubMed & 19,717 & Transductive \\
\bottomrule
\end{tabular}
\end{table}

\subsubsection{Datasets and Evaluation Protocol}
We use five standard graph-learning datasets: MUTAG, ENZYMES, PROTEINS\_full, Cora, and PubMed~\citep{morris2020tudataset,yang2016revisiting}, spanning inductive/transductive regimes and continuous/categorical node attributes. Inductive datasets support graph classification and, when node features are available, node classification; transductive datasets support node classification and LP, while inductive LP is omitted because clean baselines are weak. Inductive tasks use graph-level splits, whereas transductive node and LP tasks use node-mask splits on a shared graph. For LP, to avoid target-link leakage, we adopt the target-edge exclusion setting studied by Zhu et al.~\citep{zhu2023spottarget}: validation and test positives are removed from the shared message-passing graph and paired with equal numbers of negative targets; training positives are sampled from real edges whose endpoints both lie in the training node partition and remain available during training and perturbation optimization. Validation and test positives are sampled from the remaining real edges, creating a stricter evaluation scenario in which the learner must predict unseen links that may involve either perturbed training nodes or unperturbed nodes outside that partition.


\subsubsection{Model Architectures and Training Frameworks}
We evaluate four backbones: GCN~\citep{kipf2017semi}, GAT~\citep{velickovic2018graph}, GIN~\citep{xu2019powerful}, and GraphSAGE~\citep{hamilton2017inductive}. We additionally test the perturbations under self-supervised learning frameworks (GraphMAE~\citep{hou2022graphmae} and GRACE~\citep{zhu2020deep}).

\subsubsection{Evaluation Metrics and Baselines}
For the primary evaluations, we report Macro-F1 for graph classification, F1 for node classification, and AUC for link prediction; graph accuracy is additionally reported in selected robustness and ablation studies. We compare against clean training and budget-matched random perturbations: variance-matched Gaussian noise for continuous attributes and random categorical flips for discrete attributes. We summarize degradation using \(\Delta_c=m_{\mathrm{apply}}-m_{\mathrm{no}}\) and \(\Delta_r=m_{\mathrm{apply}}-m_{\mathrm{random}}\); negative gaps indicate degradation, with consistently negative gaps providing stronger evidence of unlearnability.
Perturbation budgets follow Section~\ref{subsec:perturbation-parameterization}; unless otherwise stated, we set \(\rho=0.2\) for continuous attributes and \(\beta=0.05\) for categorical attributes.
\subsection{Effectiveness and Transferability}
\providecommand{\meanstd}[2]{#1{$\pm$#2}}
\subsubsection{Within-backend effectiveness}
Across tasks, optimized perturbations generally reduce clean-test performance relative to both clean training and matched-variance Gaussian noise, demonstrating strong same-backend unlearnability; the corresponding supervised results are reported in Tables~\ref{tab:main-supervised-graph}--\ref{tab:main-supervised-link}.

\begin{table}[t]
\centering
\small
\setlength{\tabcolsep}{2pt}
\renewcommand{\arraystretch}{0.92}
\caption{Same-backend supervised results for graph classification (Macro-F1). Entries are mean{$\pm$std} over three runs. Bold: \(\mu+\sigma\geq0\).}
\label{tab:main-supervised-graph}
\begin{tabular}{llccc}
\toprule
Dataset & Enc. & Clean & \(\Delta_c\) & \(\Delta_r\) \\
\midrule
MUTAG & GAT & \meanstd{0.815}{0.030} & \negnum{\meanstd{-0.300}{0.144}} & \negnum{\meanstd{-0.259}{0.162}} \\
MUTAG & GCN & \meanstd{0.883}{0.029} & \negnum{\meanstd{-0.223}{0.053}} & \negnum{\meanstd{-0.114}{0.051}} \\
MUTAG & GIN & \meanstd{0.826}{0.024} & \negnum{\meanstd{-0.338}{0.127}} & \negnum{\meanstd{-0.286}{0.135}} \\
MUTAG & SAGE & \meanstd{0.859}{0.032} & \negnum{\meanstd{-0.307}{0.111}} & \negnum{\meanstd{-0.229}{0.080}} \\
\midrule
ENZYMES & GAT & \meanstd{0.512}{0.030} & \negnum{\meanstd{-0.290}{0.036}} & \negnum{\meanstd{-0.317}{0.035}} \\
ENZYMES & GCN & \meanstd{0.512}{0.030} & \negnum{\meanstd{-0.287}{0.046}} & \negnum{\meanstd{-0.307}{0.064}} \\
ENZYMES & GIN & \meanstd{0.579}{0.065} & \negnum{\meanstd{-0.389}{0.085}} & \negnum{\meanstd{-0.417}{0.067}} \\
ENZYMES & SAGE & \meanstd{0.705}{0.025} & \negnum{\meanstd{-0.507}{0.028}} & \negnum{\meanstd{-0.524}{0.027}} \\
\midrule
PROTEINS\_full & GAT & \meanstd{0.641}{0.024} & \negnum{\meanstd{-0.153}{0.102}} & \negnum{\meanstd{-0.127}{0.083}} \\
PROTEINS\_full & GCN & \meanstd{0.675}{0.027} & \negnum{\meanstd{-0.165}{0.031}} & \negnum{\meanstd{-0.167}{0.042}} \\
PROTEINS\_full & GIN & \meanstd{0.632}{0.008} & \negnum{\meanstd{-0.263}{0.052}} & \negnum{\meanstd{-0.247}{0.076}} \\
PROTEINS\_full & SAGE & \meanstd{0.642}{0.010} & \negnum{\meanstd{-0.279}{0.067}} & \negnum{\meanstd{-0.347}{0.036}} \\
\bottomrule
\end{tabular}
\end{table}

\begin{table}[t]
\centering
\small
\setlength{\tabcolsep}{2pt}
\renewcommand{\arraystretch}{0.92}
\caption{Same-backend supervised results for node classification (F1). Entries are mean{$\pm$std} over three runs. Bold: \(\mu+\sigma\geq0\).}
\label{tab:main-supervised-node}
\begin{tabular}{llccc}
\toprule
Dataset & Enc. & Clean & \(\Delta_c\) & \(\Delta_r\) \\
\midrule
ENZYMES & GAT & \meanstd{0.737}{0.012} & \negnum{\meanstd{-0.023}{0.017}} & \negnum{\meanstd{-0.025}{0.017}} \\
ENZYMES & GCN & \meanstd{0.723}{0.008} & \negnum{\meanstd{-0.018}{0.015}} & \negnum{\meanstd{-0.017}{0.015}} \\
ENZYMES & GIN & \meanstd{0.782}{0.004} & \negnum{\meanstd{-0.014}{0.004}} & \negnum{\meanstd{-0.020}{0.004}} \\
ENZYMES & SAGE & \meanstd{0.956}{0.002} & \negnum{\meanstd{-0.015}{0.006}} & \negnum{\meanstd{-0.012}{0.006}} \\
\midrule
PROTEINS\_full & GAT & \meanstd{0.642}{0.015} & \negnum{\meanstd{-0.138}{0.027}} & \negnum{\meanstd{-0.134}{0.032}} \\
PROTEINS\_full & GCN & \meanstd{0.657}{0.009} & \negnum{\meanstd{-0.028}{0.019}} & \negnum{\meanstd{-0.024}{0.024}} \\
PROTEINS\_full & GIN & \meanstd{0.712}{0.005} & \negnum{\meanstd{-0.061}{0.027}} & \negnum{\meanstd{-0.055}{0.036}} \\
PROTEINS\_full & SAGE & \meanstd{0.860}{0.035} & \negnum{\meanstd{-0.203}{0.104}} & \negnum{\meanstd{-0.155}{0.066}} \\
\midrule
Cora & GAT & \meanstd{0.876}{0.014} & \negnum{\meanstd{-0.080}{0.021}} & \negnum{\meanstd{-0.078}{0.013}} \\
Cora & GCN & \meanstd{0.864}{0.013} & \negnum{\meanstd{-0.056}{0.028}} & \negnum{\meanstd{-0.048}{0.022}} \\
Cora & GIN & \meanstd{0.869}{0.009} & \negnum{\meanstd{-0.059}{0.030}} & \negnum{\meanstd{-0.047}{0.020}} \\
Cora & SAGE & \meanstd{0.868}{0.010} & \negnum{\meanstd{-0.079}{0.011}} & \negnum{\meanstd{-0.069}{0.012}} \\
\midrule
PubMed & GAT & \meanstd{0.889}{0.002} & \negnum{\meanstd{-0.248}{0.031}} & \negnum{\meanstd{-0.249}{0.034}} \\
PubMed & GCN & \meanstd{0.901}{0.002} & \negnum{\meanstd{-0.240}{0.018}} & \negnum{\meanstd{-0.235}{0.018}} \\
PubMed & GIN & \meanstd{0.897}{0.001} & \negnum{\meanstd{-0.253}{0.018}} & \negnum{\meanstd{-0.255}{0.019}} \\
PubMed & SAGE & \meanstd{0.900}{0.000} & \negnum{\meanstd{-0.195}{0.004}} & \negnum{\meanstd{-0.196}{0.004}} \\
\bottomrule
\end{tabular}
\end{table}

\begin{table}[t]
\centering
\small
\setlength{\tabcolsep}{2pt}
\renewcommand{\arraystretch}{0.92}
\caption{Same-backend supervised results for link prediction (AUC). Entries are mean{$\pm$std} over three runs. Bold: \(\mu+\sigma\geq0\).}
\label{tab:main-supervised-link}
\begin{tabular}{llccc}
\toprule
Dataset & Enc. & Clean & \(\Delta_c\) & \(\Delta_r\) \\
\midrule
Cora & GAT & \meanstd{0.950}{0.005} & \negnum{\meanstd{-0.066}{0.013}} & \negnum{\meanstd{-0.065}{0.013}} \\
Cora & GCN & \meanstd{0.835}{0.026} & \negnum{\meanstd{-0.079}{0.017}} & \negnum{\meanstd{-0.077}{0.035}} \\
Cora & GIN & \meanstd{0.893}{0.011} & \negnum{\meanstd{-0.143}{0.051}} & \negnum{\meanstd{-0.127}{0.038}} \\
Cora & SAGE & \meanstd{0.942}{0.005} & \negnum{\meanstd{-0.062}{0.015}} & \negnum{\meanstd{-0.059}{0.015}} \\
\midrule
PubMed & GAT & \meanstd{0.964}{0.001} & \negnum{\meanstd{-0.012}{0.003}} & \negnum{\meanstd{-0.011}{0.003}} \\
PubMed & GCN & \meanstd{0.970}{0.002} & \negnum{\meanstd{-0.012}{0.005}} & \negnum{\meanstd{-0.015}{0.006}} \\
PubMed & GIN & \meanstd{0.916}{0.002} & \textbf{\meanstd{-0.022}{0.023}} & \negnum{\meanstd{-0.029}{0.021}} \\
PubMed & SAGE & \meanstd{0.971}{0.001} & \negnum{\meanstd{-0.015}{0.003}} & \negnum{\meanstd{-0.015}{0.003}} \\
\bottomrule
\end{tabular}
\end{table}

\subsubsection{Cross-backend effectiveness}
\begin{figure*}[t]
\centering
\includegraphics[width=\textwidth]{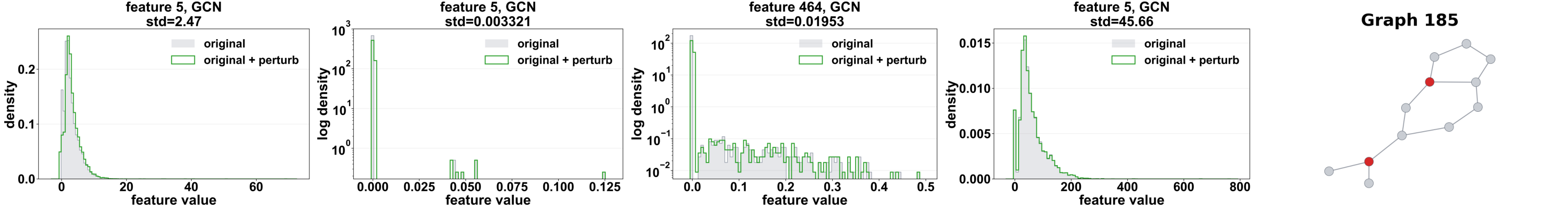}
\caption{Representative GCN perturbations on five datasets. Red nodes mark category flips in the MUTAG panel; Cora and PubMed use log-scaled density axes for skewed distributions.}
\Description{Five dataset panels visualize representative GCN-generated feature perturbations. Continuous-feature panels compare original and perturbed distributions; the MUTAG panel marks nodes whose categories changed in red. Cora and PubMed density plots use logarithmic vertical axes.}
\label{fig:gcn-perturbation-overview}
\end{figure*}

Table~\ref{tab:main-transfer} reports cross-backend gaps, showing that perturbations retain their degrading effects across most tasks. We attribute this transferability to our Task-Aligned Separability Objective (TASO) and Type-Adaptive Perturbation (TAP), whose contributions are analyzed in Section~\ref{subsec:ablation}. Figure~\ref{fig:gcn-perturbation-overview} shows the distributions of clean and perturbed data.

\begin{table*}[t]
\centering
\caption{Cross-backend gaps against clean training (\(\Delta_c\)) and matched random-perturbation training (\(\Delta_r\)). Task metrics are graph Macro-F1, node F1, and link AUC. Rows and columns denote source and target backbones; entries are mean{$\pm$std} over three seeds. Bold: \(\mu+\sigma\geq0\).}
\label{tab:main-transfer}
\resizebox{\textwidth}{!}{%
\begin{tabular}{lllcccccccc}
\toprule
 & & & \multicolumn{4}{c}{\(\Delta_c\)} & \multicolumn{4}{c}{\(\Delta_r\)} \\
\cmidrule(lr){4-7} \cmidrule(lr){8-11}
Dataset & Task & Source & GAT & GCN & GIN & SAGE & GAT & GCN & GIN & SAGE \\
\midrule
\multirow{4}{*}{MUTAG} & \multirow{4}{*}{Graph} & GAT & \negnum{\meanstd{-0.300}{0.144}} & \negnum{\meanstd{-0.279}{0.142}} & \negnum{\meanstd{-0.194}{0.074}} & \negnum{\meanstd{-0.276}{0.196}} & \negnum{\meanstd{-0.259}{0.162}} & \negnum{\meanstd{-0.170}{0.146}} & \negnum{\meanstd{-0.142}{0.091}} & \textbf{\meanstd{-0.198}{0.203}} \\
 &  & GCN & \negnum{\meanstd{-0.268}{0.057}} & \negnum{\meanstd{-0.223}{0.053}} & \negnum{\meanstd{-0.240}{0.118}} & \negnum{\meanstd{-0.255}{0.041}} & \negnum{\meanstd{-0.227}{0.107}} & \negnum{\meanstd{-0.114}{0.051}} & \negnum{\meanstd{-0.188}{0.079}} & \negnum{\meanstd{-0.177}{0.073}} \\
 &  & GIN & \negnum{\meanstd{-0.275}{0.252}} & \negnum{\meanstd{-0.244}{0.133}} & \negnum{\meanstd{-0.338}{0.127}} & \negnum{\meanstd{-0.408}{0.184}} & \negnum{\meanstd{-0.234}{0.206}} & \negnum{\meanstd{-0.135}{0.132}} & \negnum{\meanstd{-0.286}{0.135}} & \negnum{\meanstd{-0.330}{0.166}} \\
 &  & SAGE & \negnum{\meanstd{-0.252}{0.069}} & \negnum{\meanstd{-0.338}{0.101}} & \negnum{\meanstd{-0.230}{0.043}} & \negnum{\meanstd{-0.307}{0.111}} & \negnum{\meanstd{-0.211}{0.098}} & \negnum{\meanstd{-0.230}{0.097}} & \negnum{\meanstd{-0.178}{0.060}} & \negnum{\meanstd{-0.229}{0.080}} \\
\midrule
\multirow{8}{*}{ENZYMES} & \multirow{4}{*}{Graph} & GAT & \negnum{\meanstd{-0.290}{0.036}} & \negnum{\meanstd{-0.252}{0.031}} & \negnum{\meanstd{-0.349}{0.098}} & \negnum{\meanstd{-0.511}{0.037}} & \negnum{\meanstd{-0.317}{0.035}} & \negnum{\meanstd{-0.254}{0.081}} & \negnum{\meanstd{-0.377}{0.049}} & \negnum{\meanstd{-0.534}{0.038}} \\
 &  & GCN & \negnum{\meanstd{-0.272}{0.180}} & \negnum{\meanstd{-0.287}{0.046}} & \negnum{\meanstd{-0.383}{0.056}} & \negnum{\meanstd{-0.521}{0.036}} & \negnum{\meanstd{-0.298}{0.173}} & \negnum{\meanstd{-0.307}{0.064}} & \negnum{\meanstd{-0.415}{0.066}} & \negnum{\meanstd{-0.531}{0.029}} \\
 &  & GIN & \negnum{\meanstd{-0.368}{0.043}} & \negnum{\meanstd{-0.285}{0.023}} & \negnum{\meanstd{-0.389}{0.085}} & \negnum{\meanstd{-0.550}{0.062}} & \negnum{\meanstd{-0.365}{0.065}} & \negnum{\meanstd{-0.252}{0.007}} & \negnum{\meanstd{-0.417}{0.067}} & \negnum{\meanstd{-0.549}{0.098}} \\
 &  & SAGE & \negnum{\meanstd{-0.291}{0.113}} & \negnum{\meanstd{-0.263}{0.041}} & \negnum{\meanstd{-0.410}{0.070}} & \negnum{\meanstd{-0.507}{0.028}} & \negnum{\meanstd{-0.325}{0.125}} & \negnum{\meanstd{-0.239}{0.050}} & \negnum{\meanstd{-0.448}{0.058}} & \negnum{\meanstd{-0.524}{0.027}} \\
\cmidrule(lr){2-11}
 & \multirow{4}{*}{Node} & GAT & \negnum{\meanstd{-0.023}{0.017}} & \negnum{\meanstd{-0.016}{0.012}} & \textbf{\meanstd{-0.012}{0.019}} & \negnum{\meanstd{-0.013}{0.003}} & \negnum{\meanstd{-0.025}{0.017}} & \negnum{\meanstd{-0.018}{0.015}} & \negnum{\meanstd{-0.021}{0.020}} & \negnum{\meanstd{-0.009}{0.001}} \\
 &  & GCN & \negnum{\meanstd{-0.022}{0.018}} & \negnum{\meanstd{-0.018}{0.015}} & \negnum{\meanstd{-0.019}{0.015}} & \negnum{\meanstd{-0.013}{0.003}} & \negnum{\meanstd{-0.025}{0.020}} & \negnum{\meanstd{-0.017}{0.015}} & \negnum{\meanstd{-0.024}{0.016}} & \negnum{\meanstd{-0.009}{0.004}} \\
 &  & GIN & \negnum{\meanstd{-0.021}{0.015}} & \negnum{\meanstd{-0.021}{0.012}} & \negnum{\meanstd{-0.014}{0.004}} & \negnum{\meanstd{-0.011}{0.000}} & \negnum{\meanstd{-0.020}{0.017}} & \negnum{\meanstd{-0.021}{0.018}} & \negnum{\meanstd{-0.020}{0.004}} & \negnum{\meanstd{-0.008}{0.001}} \\
 &  & SAGE & \negnum{\meanstd{-0.009}{0.005}} & \textbf{\meanstd{-0.005}{0.005}} & \negnum{\meanstd{-0.012}{0.011}} & \negnum{\meanstd{-0.015}{0.006}} & \textbf{\meanstd{-0.011}{0.014}} & \textbf{\meanstd{-0.002}{0.005}} & \negnum{\meanstd{-0.016}{0.011}} & \negnum{\meanstd{-0.012}{0.006}} \\
\midrule
\multirow{8}{*}{PROTEINS} & \multirow{4}{*}{Graph} & GAT & \negnum{\meanstd{-0.153}{0.102}} & \negnum{\meanstd{-0.092}{0.083}} & \textbf{\meanstd{-0.089}{0.127}} & \negnum{\meanstd{-0.079}{0.069}} & \negnum{\meanstd{-0.127}{0.083}} & \negnum{\meanstd{-0.117}{0.082}} & \textbf{\meanstd{-0.083}{0.142}} & \negnum{\meanstd{-0.155}{0.115}} \\
 &  & GCN & \negnum{\meanstd{-0.317}{0.052}} & \negnum{\meanstd{-0.165}{0.031}} & \negnum{\meanstd{-0.212}{0.086}} & \negnum{\meanstd{-0.252}{0.118}} & \negnum{\meanstd{-0.274}{0.063}} & \negnum{\meanstd{-0.167}{0.042}} & \negnum{\meanstd{-0.282}{0.050}} & \negnum{\meanstd{-0.318}{0.071}} \\
 &  & GIN & \negnum{\meanstd{-0.322}{0.023}} & \negnum{\meanstd{-0.192}{0.035}} & \negnum{\meanstd{-0.263}{0.052}} & \negnum{\meanstd{-0.355}{0.017}} & \negnum{\meanstd{-0.257}{0.031}} & \negnum{\meanstd{-0.200}{0.032}} & \negnum{\meanstd{-0.247}{0.076}} & \negnum{\meanstd{-0.426}{0.020}} \\
 &  & SAGE & \negnum{\meanstd{-0.161}{0.111}} & \textbf{\meanstd{-0.035}{0.104}} & \negnum{\meanstd{-0.205}{0.058}} & \negnum{\meanstd{-0.279}{0.067}} & \negnum{\meanstd{-0.168}{0.160}} & \textbf{\meanstd{-0.019}{0.127}} & \negnum{\meanstd{-0.264}{0.102}} & \negnum{\meanstd{-0.347}{0.036}} \\
\cmidrule(lr){2-11}
 & \multirow{4}{*}{Node} & GAT & \negnum{\meanstd{-0.138}{0.027}} & \negnum{\meanstd{-0.081}{0.026}} & \negnum{\meanstd{-0.103}{0.043}} & \negnum{\meanstd{-0.462}{0.200}} & \negnum{\meanstd{-0.134}{0.032}} & \negnum{\meanstd{-0.074}{0.024}} & \negnum{\meanstd{-0.099}{0.046}} & \negnum{\meanstd{-0.426}{0.172}} \\
 &  & GCN & \negnum{\meanstd{-0.133}{0.026}} & \negnum{\meanstd{-0.028}{0.019}} & \negnum{\meanstd{-0.063}{0.018}} & \negnum{\meanstd{-0.311}{0.019}} & \negnum{\meanstd{-0.134}{0.020}} & \negnum{\meanstd{-0.024}{0.024}} & \negnum{\meanstd{-0.062}{0.021}} & \negnum{\meanstd{-0.266}{0.024}} \\
 &  & GIN & \negnum{\meanstd{-0.113}{0.028}} & \negnum{\meanstd{-0.047}{0.010}} & \negnum{\meanstd{-0.061}{0.027}} & \negnum{\meanstd{-0.288}{0.050}} & \negnum{\meanstd{-0.116}{0.015}} & \negnum{\meanstd{-0.039}{0.014}} & \negnum{\meanstd{-0.055}{0.036}} & \negnum{\meanstd{-0.247}{0.026}} \\
 &  & SAGE & \negnum{\meanstd{-0.052}{0.019}} & \negnum{\meanstd{-0.026}{0.005}} & \negnum{\meanstd{-0.040}{0.020}} & \negnum{\meanstd{-0.203}{0.104}} & \negnum{\meanstd{-0.051}{0.028}} & \negnum{\meanstd{-0.020}{0.006}} & \negnum{\meanstd{-0.037}{0.025}} & \negnum{\meanstd{-0.155}{0.066}} \\
\midrule
\multirow{8}{*}{Cora} & \multirow{4}{*}{Node} & GAT & \negnum{\meanstd{-0.080}{0.021}} & \negnum{\meanstd{-0.065}{0.020}} & \negnum{\meanstd{-0.056}{0.004}} & \negnum{\meanstd{-0.058}{0.015}} & \negnum{\meanstd{-0.078}{0.013}} & \negnum{\meanstd{-0.059}{0.040}} & \negnum{\meanstd{-0.044}{0.013}} & \negnum{\meanstd{-0.049}{0.017}} \\
 &  & GCN & \negnum{\meanstd{-0.072}{0.019}} & \negnum{\meanstd{-0.056}{0.028}} & \negnum{\meanstd{-0.077}{0.022}} & \negnum{\meanstd{-0.090}{0.006}} & \negnum{\meanstd{-0.066}{0.019}} & \negnum{\meanstd{-0.048}{0.022}} & \negnum{\meanstd{-0.072}{0.026}} & \negnum{\meanstd{-0.085}{0.006}} \\
 &  & GIN & \negnum{\meanstd{-0.031}{0.006}} & \negnum{\meanstd{-0.049}{0.015}} & \negnum{\meanstd{-0.059}{0.030}} & \negnum{\meanstd{-0.038}{0.006}} & \negnum{\meanstd{-0.023}{0.006}} & \negnum{\meanstd{-0.048}{0.023}} & \negnum{\meanstd{-0.047}{0.020}} & \negnum{\meanstd{-0.031}{0.015}} \\
 &  & SAGE & \negnum{\meanstd{-0.048}{0.022}} & \negnum{\meanstd{-0.045}{0.032}} & \negnum{\meanstd{-0.056}{0.013}} & \negnum{\meanstd{-0.079}{0.011}} & \negnum{\meanstd{-0.043}{0.015}} & \negnum{\meanstd{-0.036}{0.021}} & \negnum{\meanstd{-0.050}{0.008}} & \negnum{\meanstd{-0.069}{0.012}} \\
\cmidrule(lr){2-11}
 & \multirow{4}{*}{Link} & GAT & \negnum{\meanstd{-0.066}{0.013}} & \textbf{\meanstd{-0.047}{0.092}} & \negnum{\meanstd{-0.070}{0.029}} & \negnum{\meanstd{-0.049}{0.009}} & \negnum{\meanstd{-0.065}{0.013}} & \textbf{\meanstd{-0.047}{0.084}} & \negnum{\meanstd{-0.057}{0.024}} & \negnum{\meanstd{-0.047}{0.010}} \\
 &  & GCN & \negnum{\meanstd{-0.110}{0.011}} & \negnum{\meanstd{-0.079}{0.017}} & \negnum{\meanstd{-0.073}{0.019}} & \negnum{\meanstd{-0.053}{0.016}} & \negnum{\meanstd{-0.103}{0.021}} & \negnum{\meanstd{-0.077}{0.035}} & \negnum{\meanstd{-0.057}{0.015}} & \negnum{\meanstd{-0.050}{0.018}} \\
 &  & GIN & \negnum{\meanstd{-0.083}{0.015}} & \negnum{\meanstd{-0.110}{0.045}} & \negnum{\meanstd{-0.143}{0.051}} & \negnum{\meanstd{-0.046}{0.015}} & \negnum{\meanstd{-0.083}{0.016}} & \negnum{\meanstd{-0.120}{0.021}} & \negnum{\meanstd{-0.127}{0.038}} & \negnum{\meanstd{-0.044}{0.016}} \\
 &  & SAGE & \negnum{\meanstd{-0.032}{0.014}} & \negnum{\meanstd{-0.055}{0.023}} & \negnum{\meanstd{-0.051}{0.030}} & \negnum{\meanstd{-0.062}{0.015}} & \negnum{\meanstd{-0.030}{0.018}} & \negnum{\meanstd{-0.060}{0.019}} & \negnum{\meanstd{-0.051}{0.027}} & \negnum{\meanstd{-0.059}{0.015}} \\
\midrule
\multirow{8}{*}{PubMed} & \multirow{4}{*}{Node} & GAT & \negnum{\meanstd{-0.248}{0.031}} & \negnum{\meanstd{-0.192}{0.011}} & \negnum{\meanstd{-0.222}{0.014}} & \negnum{\meanstd{-0.232}{0.013}} & \negnum{\meanstd{-0.249}{0.034}} & \negnum{\meanstd{-0.187}{0.016}} & \negnum{\meanstd{-0.222}{0.011}} & \negnum{\meanstd{-0.233}{0.015}} \\
 &  & GCN & \negnum{\meanstd{-0.156}{0.022}} & \negnum{\meanstd{-0.240}{0.018}} & \negnum{\meanstd{-0.229}{0.027}} & \negnum{\meanstd{-0.212}{0.026}} & \negnum{\meanstd{-0.157}{0.025}} & \negnum{\meanstd{-0.235}{0.018}} & \negnum{\meanstd{-0.228}{0.026}} & \negnum{\meanstd{-0.212}{0.029}} \\
 &  & GIN & \negnum{\meanstd{-0.184}{0.059}} & \negnum{\meanstd{-0.221}{0.029}} & \negnum{\meanstd{-0.253}{0.018}} & \negnum{\meanstd{-0.199}{0.005}} & \negnum{\meanstd{-0.184}{0.065}} & \negnum{\meanstd{-0.217}{0.032}} & \negnum{\meanstd{-0.255}{0.019}} & \negnum{\meanstd{-0.202}{0.005}} \\
 &  & SAGE & \negnum{\meanstd{-0.157}{0.012}} & \negnum{\meanstd{-0.149}{0.008}} & \negnum{\meanstd{-0.158}{0.011}} & \negnum{\meanstd{-0.195}{0.004}} & \negnum{\meanstd{-0.153}{0.016}} & \negnum{\meanstd{-0.145}{0.010}} & \negnum{\meanstd{-0.158}{0.010}} & \negnum{\meanstd{-0.196}{0.004}} \\
\cmidrule(lr){2-11}
 & \multirow{4}{*}{Link} & GAT & \negnum{\meanstd{-0.012}{0.003}} & \negnum{\meanstd{-0.013}{0.006}} & \negnum{\meanstd{-0.050}{0.016}} & \negnum{\meanstd{-0.019}{0.001}} & \negnum{\meanstd{-0.011}{0.003}} & \negnum{\meanstd{-0.018}{0.003}} & \negnum{\meanstd{-0.054}{0.013}} & \negnum{\meanstd{-0.019}{0.001}} \\
 &  & GCN & \negnum{\meanstd{-0.008}{0.001}} & \negnum{\meanstd{-0.012}{0.005}} & \negnum{\meanstd{-0.026}{0.017}} & \negnum{\meanstd{-0.020}{0.002}} & \negnum{\meanstd{-0.006}{0.001}} & \negnum{\meanstd{-0.015}{0.006}} & \negnum{\meanstd{-0.034}{0.019}} & \negnum{\meanstd{-0.020}{0.001}} \\
 &  & GIN & \negnum{\meanstd{-0.010}{0.001}} & \negnum{\meanstd{-0.012}{0.003}} & \textbf{\meanstd{-0.022}{0.023}} & \negnum{\meanstd{-0.024}{0.001}} & \negnum{\meanstd{-0.009}{0.002}} & \negnum{\meanstd{-0.015}{0.006}} & \negnum{\meanstd{-0.029}{0.021}} & \negnum{\meanstd{-0.024}{0.002}} \\
 &  & SAGE & \negnum{\meanstd{-0.008}{0.001}} & \negnum{\meanstd{-0.004}{0.003}} & \negnum{\meanstd{-0.029}{0.013}} & \negnum{\meanstd{-0.015}{0.003}} & \negnum{\meanstd{-0.007}{0.001}} & \negnum{\meanstd{-0.010}{0.002}} & \negnum{\meanstd{-0.035}{0.014}} & \negnum{\meanstd{-0.015}{0.003}} \\
\bottomrule
\end{tabular}
}
\end{table*}
\subsubsection{Apply on Self-Supervised Learning Framework}
We further apply perturbations optimized with the supervised framework to GraphMAE masked-feature reconstruction and GRACE augmentation-based contrastive learning~\citep{hou2022graphmae,zhu2020deep}. Although GraphMAE's feature masking and reconstruction and GRACE's stochastic views and contrastive alignment can disrupt supervised perturbation patterns, the perturbations continue to degrade downstream performance under both frameworks (Table~\ref{tab:ssl-from-supervised-proteins-full}). For GRACE, we follow the original implementation and use a GCN encoder.

\begin{table}[t]
\centering
\caption{Transfer of supervised perturbations to self-supervised learning on PROTEINS\_full. Node and graph results report F1 and Macro-F1, respectively. Entries are mean{$\pm$std} over three runs with different random seeds.}
\label{tab:ssl-from-supervised-proteins-full}
\resizebox{\columnwidth}{!}{
\begin{tabular}{llccc}
\toprule
\multicolumn{5}{c}{Node F1} \\
\cmidrule(lr){1-5}
Framework & Encoder & Clean & \(\Delta_c\) & \(\Delta_r\) \\
\midrule
\multirow{4}{*}{GraphMAE} & GIN & \meanstd{0.603}{0.006} & \negnum{\meanstd{-0.104}{0.010}} & \negnum{\meanstd{-0.100}{0.011}} \\
 & GAT & \meanstd{0.577}{0.011} & \negnum{\meanstd{-0.078}{0.037}} & \negnum{\meanstd{-0.049}{0.036}} \\
 & GCN & \meanstd{0.600}{0.021} & \negnum{\meanstd{-0.097}{0.002}} & \negnum{\meanstd{-0.093}{0.012}} \\
 & SAGE & \meanstd{0.606}{0.026} & \negnum{\meanstd{-0.092}{0.026}} & \negnum{\meanstd{-0.066}{0.008}} \\
\cmidrule(lr){1-5}
GRACE & GCN & \meanstd{0.626}{0.018} & \negnum{\meanstd{-0.122}{0.009}} & \negnum{\meanstd{-0.057}{0.036}} \\
\bottomrule
\end{tabular}
}
\par\medskip
\resizebox{\columnwidth}{!}{
\begin{tabular}{llccc}
\toprule
\multicolumn{5}{c}{Graph Macro-F1} \\
\cmidrule(lr){1-5}
Framework & Encoder & Clean & \(\Delta_c\) & \(\Delta_r\) \\
\midrule
\multirow{4}{*}{GraphMAE} & GIN & \meanstd{0.724}{0.027} & \negnum{\meanstd{-0.118}{0.054}} & \negnum{\meanstd{-0.133}{0.041}} \\
 & GAT & \meanstd{0.759}{0.001} & \negnum{\meanstd{-0.202}{0.035}} & \negnum{\meanstd{-0.184}{0.032}} \\
 & GCN & \meanstd{0.756}{0.006} & \negnum{\meanstd{-0.170}{0.085}} & \negnum{\meanstd{-0.180}{0.085}} \\
 & SAGE & \meanstd{0.754}{0.009} & \negnum{\meanstd{-0.272}{0.029}} & \negnum{\meanstd{-0.273}{0.012}} \\
\cmidrule(lr){1-5}
GRACE & GCN & \meanstd{0.720}{0.021} & \negnum{\meanstd{-0.097}{0.041}} & \negnum{\meanstd{-0.115}{0.037}} \\
\bottomrule
\end{tabular}
}
\end{table}

\subsection{Robustness Analysis}

We evaluate robustness under feature-masking and edge-dropping data augmentations and adversarial training. MUGEN-perturbed models remain below random perturbations and do not recover clean performance. Figure~\ref{fig:robustness-proteins-f1-grid} shows the GCN case on PROTEINS\_full; all results appear in Appendix~\ref{app:complete-robustness-results}.

\begin{center}
\includegraphics[width=0.90\columnwidth]{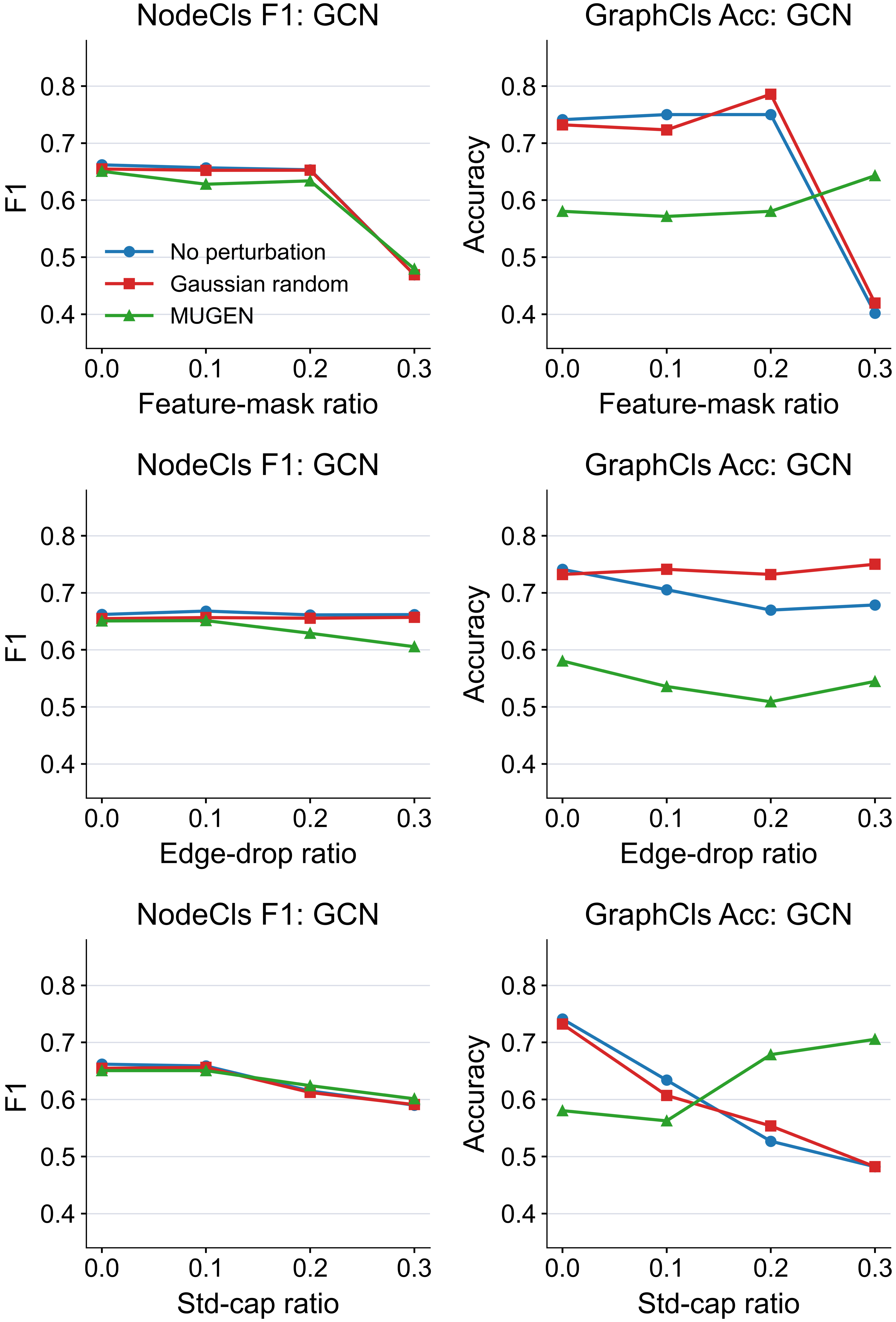}
\captionof{figure}{GCN robustness on PROTEINS\_full. Rows correspond to feature masking, edge dropping, and adversarial training; left/right panels report node F1/graph accuracy. Blue circles, red squares, and green triangles denote clean data, Gaussian random perturbations, and MUGEN perturbations.}
\Description{Six robustness plots for PROTEINS_full are arranged in three rows by intervention: feature masking, edge dropping, and adversarial training. Each row has node F1 on the left and graph accuracy on the right. Curves distinguish clean data, Gaussian random perturbations, and MUGEN perturbations using blue circles, red squares, and green triangles.}
\label{fig:robustness-proteins-f1-grid}
\end{center}

\subsection{Sensitivity to the Perturbation Budget}
\label{subsec:budget-sensitivity}
As a representative case study, we evaluate sensitivity to the continuous-feature perturbation budget on PROTEINS\_full. We set the feature-relative cap to \(\varepsilon_j=\rho\sigma_j\) and consider \(\rho\in\{0.1,0.2,0.3\}\). Even at the tightest budget, \(\rho=0.1\), GCN- and SAGE-generated perturbations yield consistently negative gaps across both tasks and all target backbones. The remaining source backbones also produce degradation, although less uniformly; increasing \(\rho\) further eliminates the remaining nonnegative gaps (Table~\ref{tab:budget-sensitivity-proteins}).

\begin{center}
\captionof{table}{Perturbation-budget sensitivity on PROTEINS\_full. Each entry reports a gap for the indicated metric, source, target, budget ratio, and baseline.}
\label{tab:budget-sensitivity-proteins}
\renewcommand{\posnum}[1]{\textbf{#1}}
\resizebox{\columnwidth}{!}{%
\begin{tabular}{cll|cccc|cccc|cccc}
\toprule
\multicolumn{3}{c|}{} & \multicolumn{4}{c|}{\(\rho=0.1\)} & \multicolumn{4}{c|}{\(\rho=0.2\)} & \multicolumn{4}{c}{\(\rho=0.3\)}\\
\cmidrule(lr){4-7}\cmidrule(lr){8-11}\cmidrule(l){12-15}
Gap & Metric & Source $\backslash$ Target & GAT & GCN & GIN & SAGE & GAT & GCN & GIN & SAGE & GAT & GCN & GIN & SAGE\\
\midrule
\multirow{8}{*}{\(\Delta_c\)} & \multirow{4}{*}{Graph acc.} & GAT & \negnum{-0.152} & \negnum{-0.232} & \negnum{-0.196} & \negnum{-0.188} & \negnum{-0.205} & \negnum{-0.170} & \negnum{-0.054} & \negnum{-0.045} & \negnum{-0.205} & \negnum{-0.152} & \negnum{-0.080} & \negnum{-0.170}\\
 &  & GCN & \negnum{-0.277} & \negnum{-0.268} & \negnum{-0.188} & \negnum{-0.304} & \negnum{-0.232} & \negnum{-0.161} & \negnum{-0.107} & \negnum{-0.241} & \negnum{-0.259} & \negnum{-0.143} & \negnum{-0.125} & \negnum{-0.214}\\
 &  & GIN & \negnum{-0.250} & \negnum{-0.250} & \negnum{-0.205} & \negnum{-0.304} & \negnum{-0.250} & \negnum{-0.179} & \negnum{-0.170} & \negnum{-0.286} & \negnum{-0.259} & \negnum{-0.250} & \negnum{-0.250} & \negnum{-0.286}\\
 &  & SAGE & \negnum{-0.045} & \negnum{-0.125} & \negnum{-0.009} & \negnum{-0.152} & \negnum{-0.134} & \posnum{+0.045} & \negnum{-0.232} & \negnum{-0.250} & \negnum{-0.223} & \negnum{-0.170} & \negnum{-0.241} & \negnum{-0.259}\\
\cmidrule(lr){2-15}
 & \multirow{4}{*}{Node F1} & GAT & \negnum{-0.138} & \negnum{-0.025} & \negnum{-0.011} & \negnum{-0.231} & \negnum{-0.139} & \negnum{-0.062} & \negnum{-0.055} & \negnum{-0.276} & \negnum{-0.148} & \negnum{-0.060} & \negnum{-0.064} & \negnum{-0.275}\\
 &  & GCN & \negnum{-0.100} & \negnum{-0.008} & \posnum{+0.005} & \negnum{-0.211} & \negnum{-0.158} & \negnum{-0.011} & \negnum{-0.047} & \negnum{-0.309} & \negnum{-0.171} & \negnum{-0.074} & \negnum{-0.070} & \negnum{-0.321}\\
 &  & GIN & \negnum{-0.161} & \negnum{-0.023} & \negnum{-0.010} & \negnum{-0.248} & \negnum{-0.144} & \negnum{-0.040} & \negnum{-0.036} & \negnum{-0.231} & \negnum{-0.156} & \negnum{-0.051} & \negnum{-0.076} & \negnum{-0.266}\\
 &  & SAGE & \negnum{-0.010} & \negnum{-0.021} & \posnum{+0.015} & \negnum{-0.207} & \negnum{-0.030} & \negnum{-0.020} & \negnum{-0.017} & \negnum{-0.140} & \negnum{-0.113} & \negnum{-0.059} & \negnum{-0.048} & \negnum{-0.221}\\
\midrule
\multirow{8}{*}{\(\Delta_r\)} & \multirow{4}{*}{Graph acc.} & GAT & \negnum{-0.098} & \negnum{-0.241} & \negnum{-0.214} & \negnum{-0.080} & \negnum{-0.152} & \negnum{-0.179} & \negnum{-0.045} & \negnum{-0.063} & \negnum{-0.063} & \negnum{-0.152} & \negnum{-0.107} & \negnum{-0.223}\\
 &  & GCN & \negnum{-0.223} & \negnum{-0.277} & \negnum{-0.205} & \negnum{-0.214} & \negnum{-0.143} & \negnum{-0.152} & \negnum{-0.205} & \negnum{-0.232} & \negnum{-0.089} & \negnum{-0.161} & \negnum{-0.170} & \negnum{-0.250}\\
 &  & GIN & \negnum{-0.188} & \negnum{-0.250} & \negnum{-0.241} & \negnum{-0.179} & \negnum{-0.152} & \negnum{-0.188} & \negnum{-0.143} & \negnum{-0.304} & \negnum{-0.089} & \negnum{-0.268} & \negnum{-0.321} & \negnum{-0.321}\\
 &  & SAGE & \posnum{+0.009} & \negnum{-0.125} & \negnum{-0.054} & \negnum{-0.054} & \negnum{-0.161} & \posnum{+0.071} & \negnum{-0.321} & \negnum{-0.250} & \negnum{-0.116} & \negnum{-0.188} & \negnum{-0.259} & \negnum{-0.286}\\
\cmidrule(lr){2-15}
 & \multirow{4}{*}{Node F1} & GAT & \negnum{-0.133} & \negnum{-0.019} & \negnum{-0.015} & \negnum{-0.255} & \negnum{-0.118} & \negnum{-0.055} & \negnum{-0.046} & \negnum{-0.265} & \negnum{-0.124} & \negnum{-0.045} & \negnum{-0.059} & \negnum{-0.245}\\
 &  & GCN & \negnum{-0.094} & \negnum{-0.001} & \posnum{+0.011} & \negnum{-0.231} & \negnum{-0.155} & \negnum{-0.004} & \negnum{-0.040} & \negnum{-0.294} & \negnum{-0.167} & \negnum{-0.063} & \negnum{-0.058} & \negnum{-0.271}\\
 &  & GIN & \negnum{-0.161} & \negnum{-0.019} & \negnum{-0.016} & \negnum{-0.270} & \negnum{-0.133} & \negnum{-0.030} & \negnum{-0.019} & \negnum{-0.220} & \negnum{-0.137} & \negnum{-0.038} & \negnum{-0.065} & \negnum{-0.225}\\
 &  & SAGE & \negnum{-0.001} & \negnum{-0.016} & \posnum{+0.019} & \negnum{-0.231} & \negnum{-0.019} & \negnum{-0.013} & \negnum{-0.008} & \negnum{-0.123} & \negnum{-0.086} & \negnum{-0.042} & \negnum{-0.030} & \negnum{-0.174}\\
\bottomrule
\end{tabular}
}
\end{center}

\subsection{Ablation}
\label{subsec:ablation}

\subsubsection{Node versus edge perturbation}
We compare node-feature or node-attribute perturbations with GradArgMax edge perturbations~\citep{liu2023graphcloak} on ENZYMES and MUTAG. The transfer results show that node perturbations yield more consistently negative transfer gaps: they are stronger and more uniform on ENZYMES and provide the more reliable transfer pattern on MUTAG despite isolated large edge-flip drops.

\begin{center}
\captionof{table}{Node- versus edge-perturbation transfer gaps (\(\Delta_c\)) for graph accuracy.}
\label{tab:appendix-node-edge-perturbation-ablation}
\resizebox{\columnwidth}{!}{
\begin{tabular}{llcccccccc}
\toprule
Dataset & Source $\backslash$ Target & \multicolumn{4}{c}{Node perturbation} & \multicolumn{4}{c}{Edge perturbation}\\
\cmidrule(lr){3-6}\cmidrule(lr){7-10}
 & & GAT & GCN & GIN & SAGE & GAT & GCN & GIN & SAGE\\
\midrule
\multirow{4}{*}{ENZYMES} & GAT & \negnum{-0.233} & \negnum{-0.233} & \negnum{-0.333} & \negnum{-0.450} & \negnum{-0.083} & \negnum{-0.100} & \negnum{-0.200} & \negnum{-0.050}\\
 & GCN & \negnum{-0.067} & \negnum{-0.300} & \negnum{-0.350} & \negnum{-0.500} & \textbf{+0.050} & \textbf{+0.000} & \negnum{-0.100} & \textbf{+0.017}\\
 & GIN & \negnum{-0.250} & \negnum{-0.233} & \negnum{-0.433} & \negnum{-0.517} & \negnum{-0.017} & \negnum{-0.067} & \negnum{-0.183} & \textbf{+0.000}\\
 & SAGE & \negnum{-0.117} & \negnum{-0.283} & \negnum{-0.450} & \negnum{-0.383} & \negnum{-0.150} & \textbf{+0.000} & \negnum{-0.333} & \negnum{-0.033}\\
\midrule
\multirow{4}{*}{MUTAG} & GAT & \negnum{-0.250} & \negnum{-0.300} & \negnum{-0.250} & \negnum{-0.300} & \negnum{-0.050} & \negnum{-0.400} & \negnum{-0.250} & \negnum{-0.100}\\
 & GCN & \negnum{-0.200} & \negnum{-0.200} & \negnum{-0.150} & \negnum{-0.200} & \textbf{+0.050} & \negnum{-0.100} & \negnum{-0.100} & \textbf{+0.050}\\
 & GIN & \textbf{+0.000} & \negnum{-0.100} & \negnum{-0.150} & \negnum{-0.200} & \textbf{+0.050} & \textbf{+0.050} & \negnum{-0.100} & \negnum{-0.150}\\
 & SAGE & \negnum{-0.200} & \negnum{-0.350} & \negnum{-0.250} & \negnum{-0.250} & \textbf{+0.050} & \negnum{-0.300} & \negnum{-0.150} & \negnum{-0.350}\\
\bottomrule
\end{tabular}
}
\end{center}

\subsubsection{Objective design for transferability}
\begin{figure}[!t]
\centering
\begin{minipage}[t]{0.32\columnwidth}
\centering
(a)
\includegraphics[width=\linewidth]{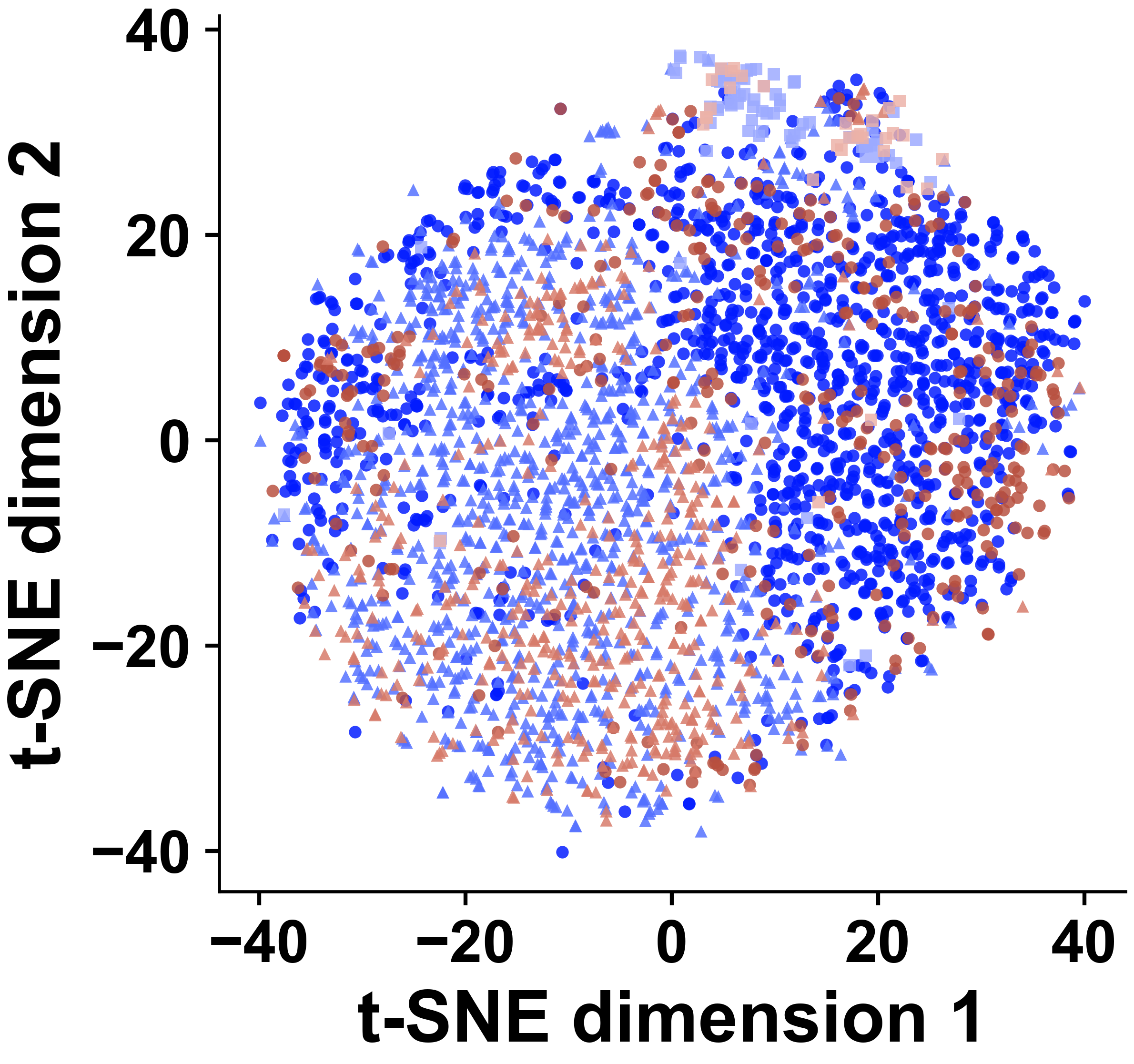}
\end{minipage}%
\hfill
\begin{minipage}[t]{0.32\columnwidth}
\centering
(b)
\includegraphics[width=\linewidth]{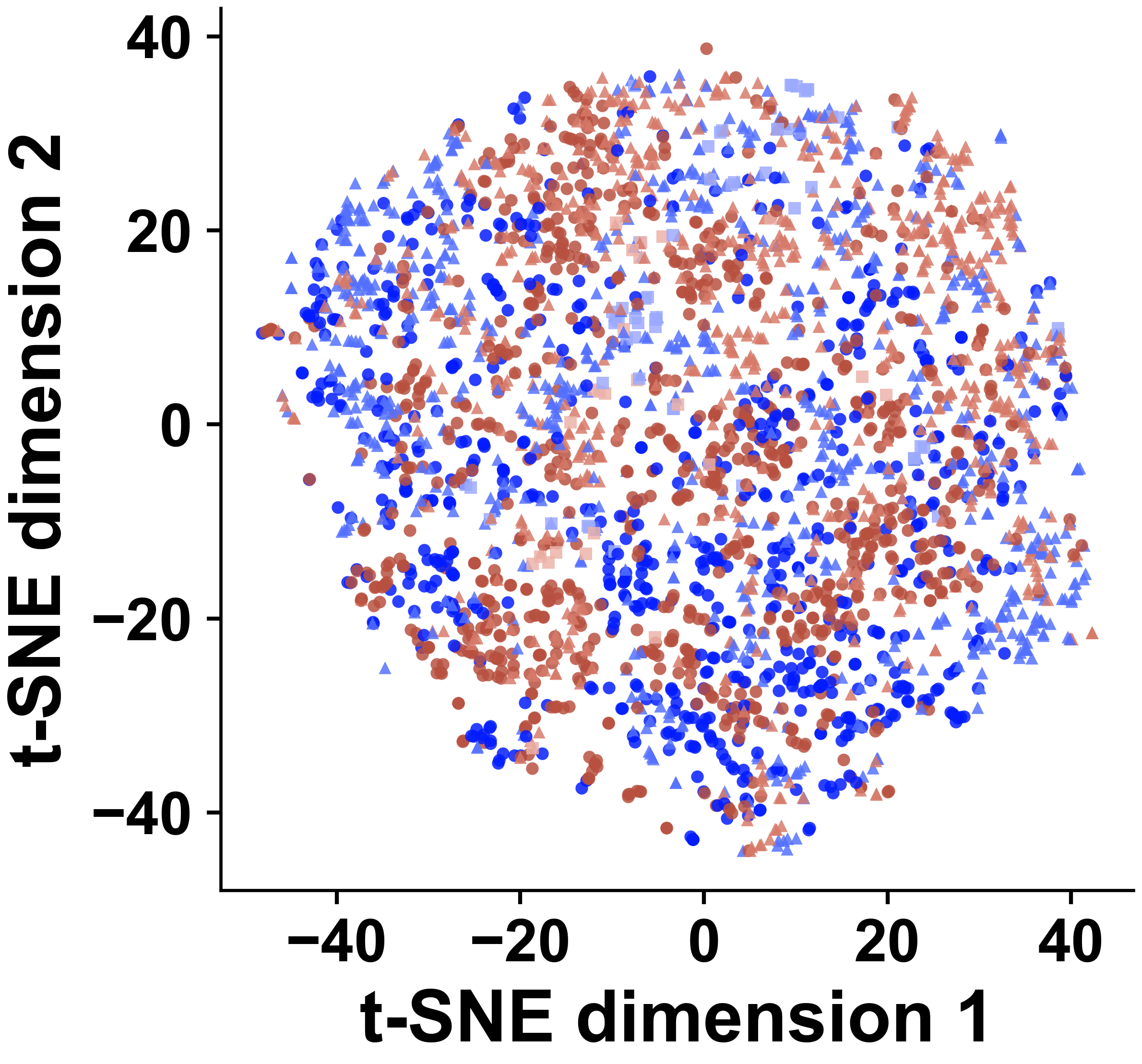}
\end{minipage}%
\hfill
\begin{minipage}[t]{0.32\columnwidth}
\centering
(c)
\includegraphics[width=\linewidth]{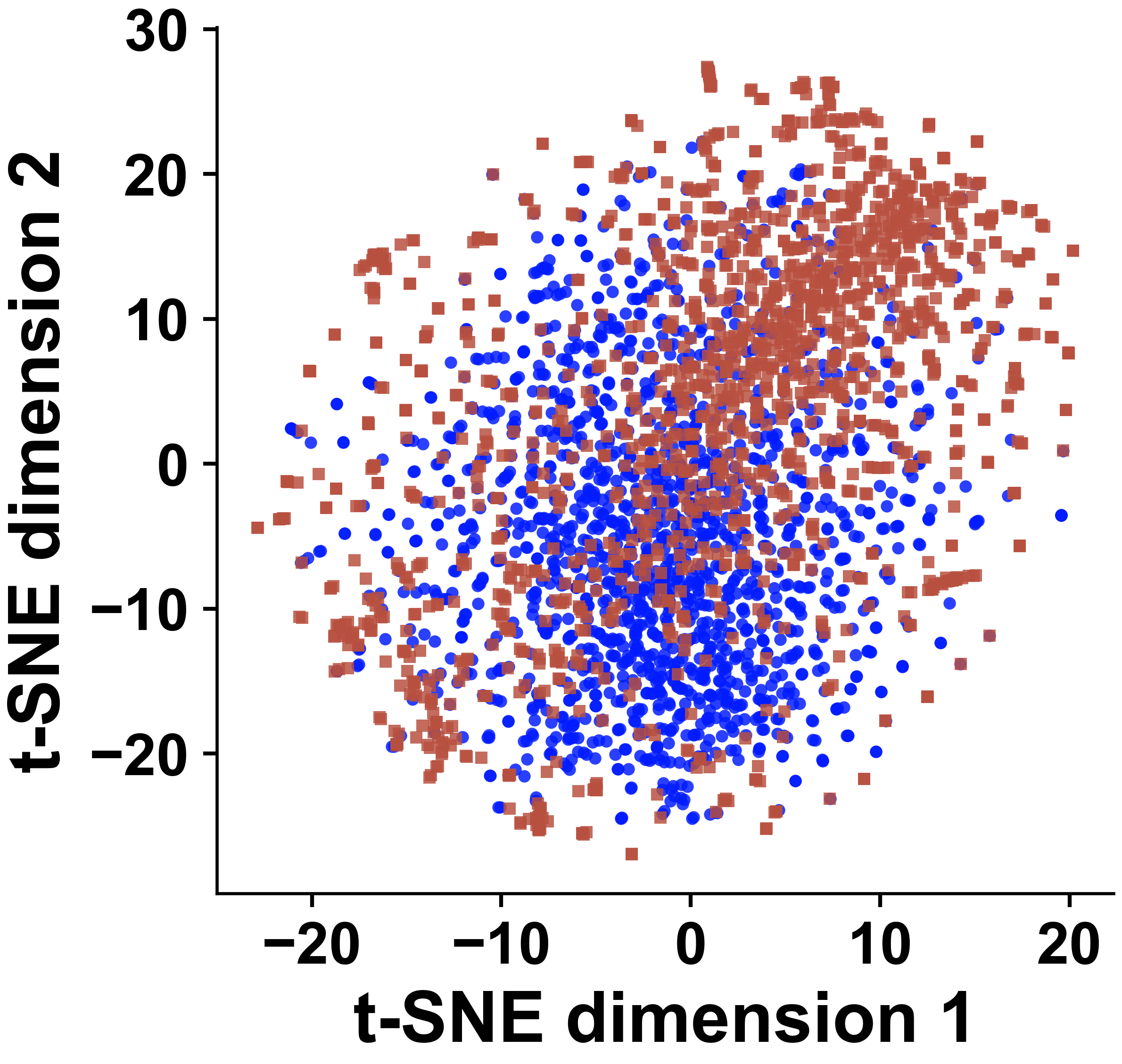}
\end{minipage}%
\par
\begin{minipage}[t]{0.32\columnwidth}
\centering
(d)
\includegraphics[width=\linewidth]{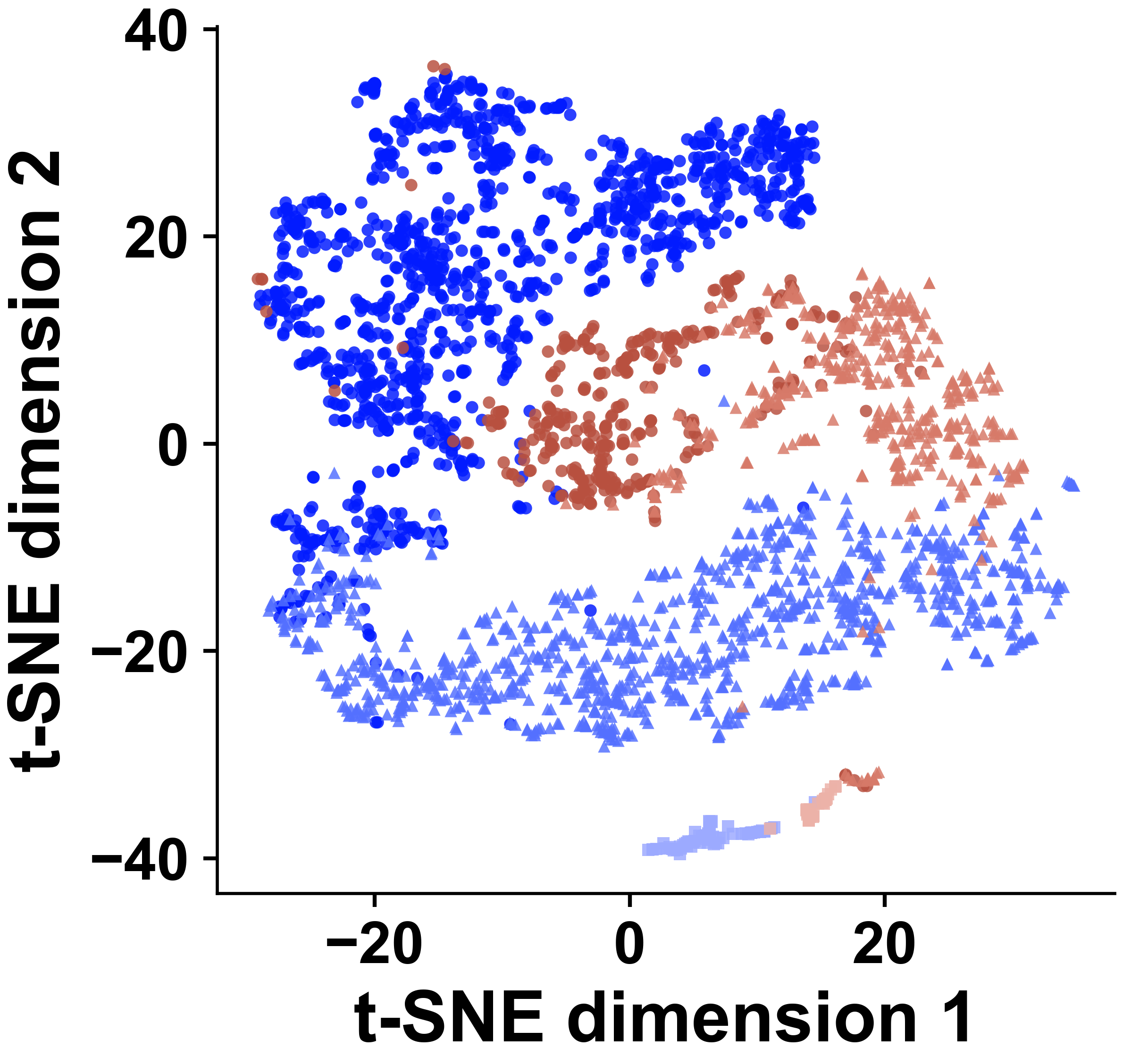}
\end{minipage}%
\hfill
\begin{minipage}[t]{0.32\columnwidth}
\centering
(e)
\includegraphics[width=\linewidth]{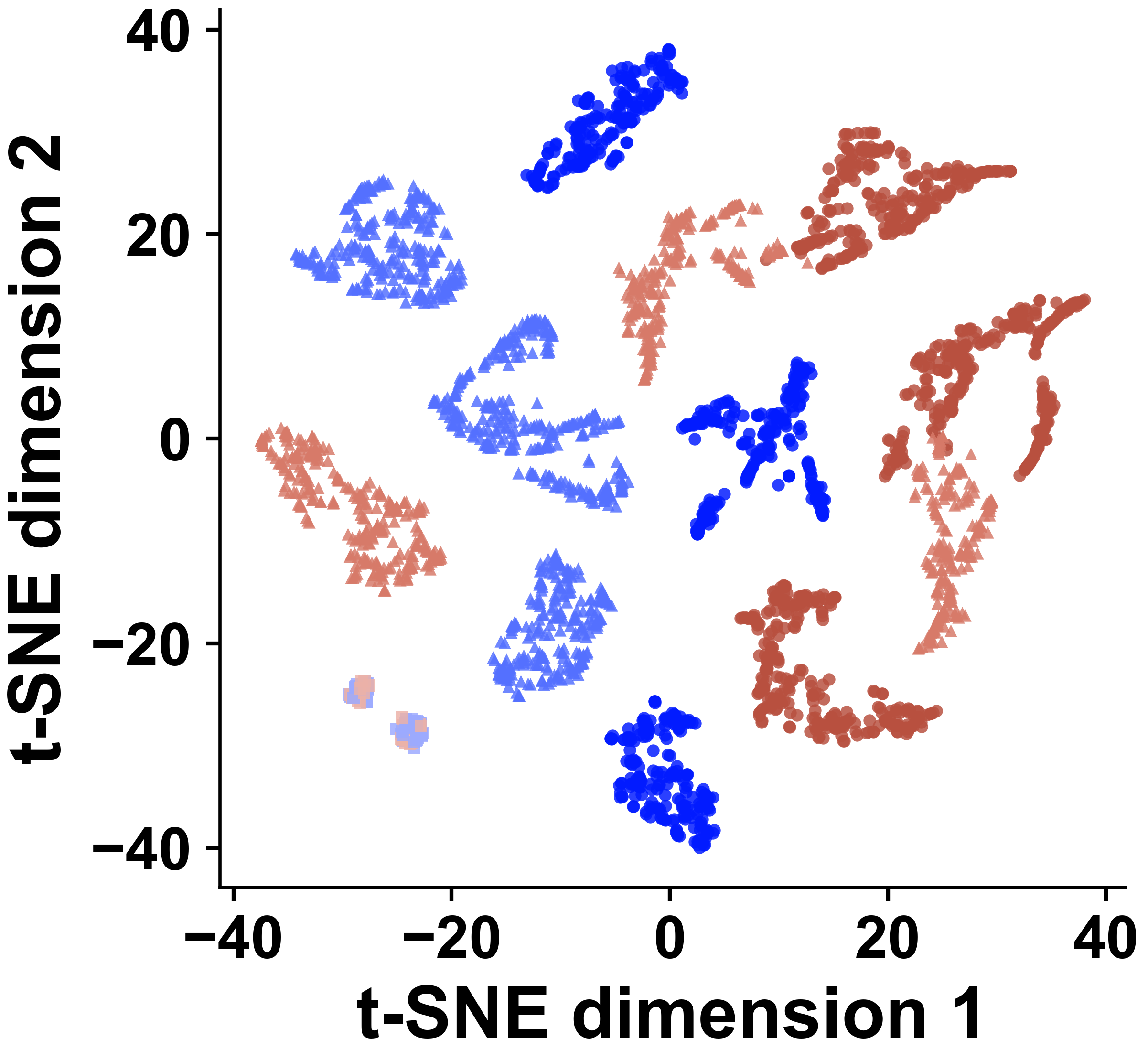}
\end{minipage}%
\hfill
\begin{minipage}[t]{0.32\columnwidth}
\centering
(f)
\includegraphics[width=\linewidth]{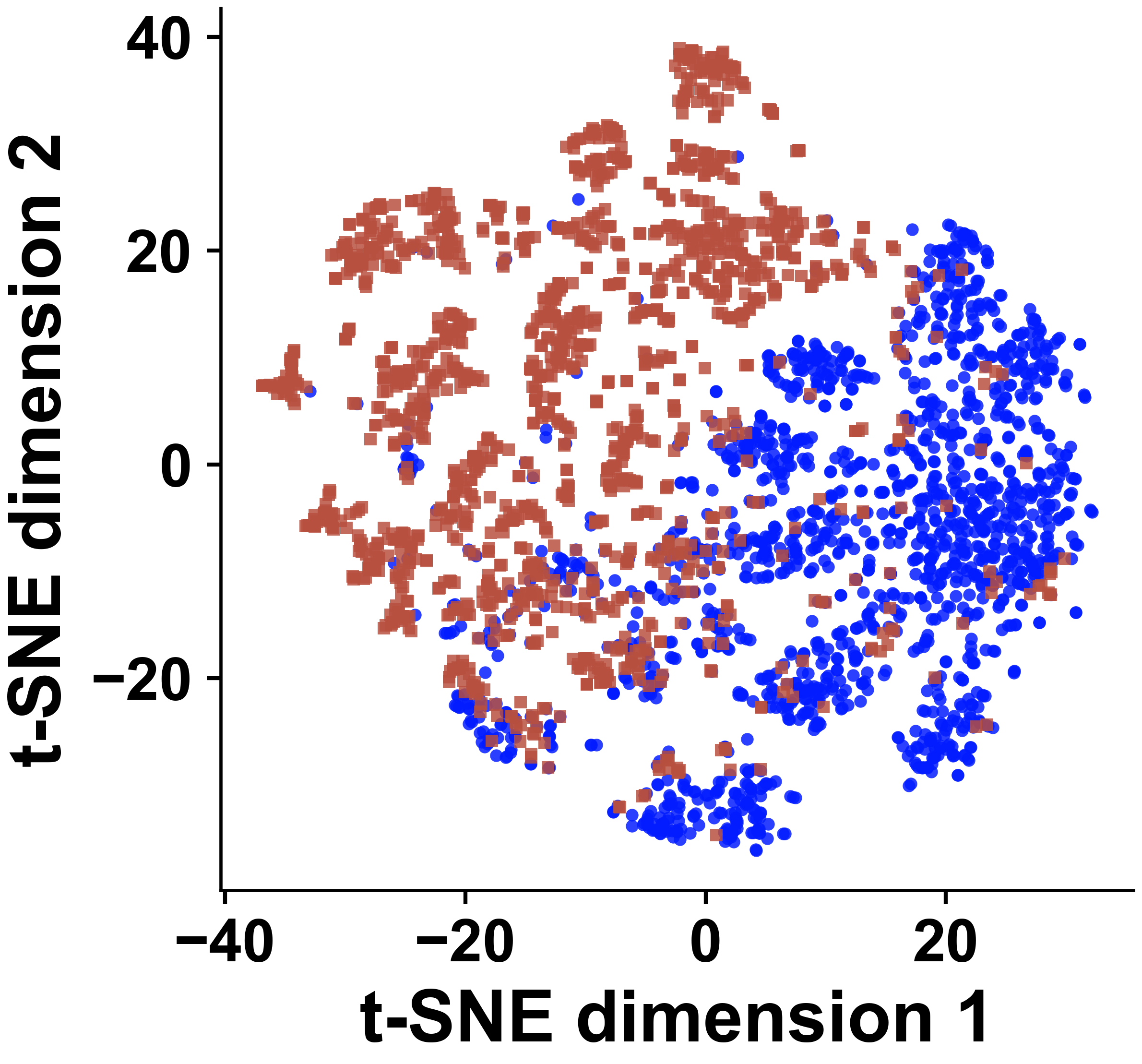}
\end{minipage}
\caption{t-SNE of standardized SAGE perturbation representations. Panels (a--c) use classification and panels (d--f) use TASO on PROTEINS\_full, ENZYMES, and Cora, respectively. Panels (c) and (f) show Hadamard link-pair features; colors and marker shapes denote LP labels.}
\Description{Six t-SNE scatter plots compare standardized SAGE perturbation representations. The top row uses classification objectives and the bottom row uses TASO for PROTEINS_full, ENZYMES, and Cora. In the Cora panels, points represent Hadamard link-pair features; color and marker shape encode link-prediction labels.}
\label{fig:objective-tsne}
\end{figure}

We compare otherwise matched supervised objectives with and without TASO under the same node-graph protocol and feature-relative budget \(\rho=0.2\) across three datasets. Figure~\ref{fig:objective-tsne} shows that TASO promotes more separable class-conditional structures. Table~\ref{tab:objective-design-diagonal} summarizes same-backbone effects, while Tables~\ref{tab:appendix-proteins-objective-design}, \ref{tab:appendix-cora-objective-design}, and~\ref{tab:enzymes-objective-design} report the full four-source-by-four-target transfer matrices for PROTEINS\_full, Cora, and ENZYMES, respectively. Across all three datasets, TASO yields stronger unlearnability and a more consistently negative transfer pattern than classification-only objectives.

\begin{table}[!htbp]
\centering
\caption{Combined objective-design ablation. Across-backend gaps relative to clean and random baselines are reported; ``--'' denotes an inapplicable task metric.}
\label{tab:objective-design-diagonal}
\renewcommand{\posnum}[1]{\textbf{#1}}
\resizebox{\columnwidth}{!}{%
\begin{tabular}{llcccc|cccc|cccc}
\toprule
\multicolumn{14}{c}{TASO} \\ 
\cmidrule(lr){1-14}
Dataset & Gap & \multicolumn{4}{c|}{Graph accuracy} & \multicolumn{4}{c|}{Node F1} & \multicolumn{4}{c}{LP AUC} \\ 
\cmidrule(lr){3-6}\cmidrule(lr){7-10}\cmidrule(lr){11-14}
 &  & GAT & GCN & GIN & SAGE & GAT & GCN & GIN & SAGE & GAT & GCN & GIN & SAGE \\ 
\midrule
\multirow{2}{*}{PROTEINS\_full} & \(\Delta_c\) & \negnum{-0.205} & \negnum{-0.161} & \negnum{-0.170} & \negnum{-0.250} & \negnum{-0.139} & \negnum{-0.011} & \negnum{-0.036} & \negnum{-0.140} & - & - & - & - \\ 
 & \(\Delta_r\) & \negnum{-0.152} & \negnum{-0.152} & \negnum{-0.143} & \negnum{-0.250} & \negnum{-0.118} & \negnum{-0.004} & \negnum{-0.019} & \negnum{-0.123} & - & - & - & - \\ 
\multirow{2}{*}{Cora} & \(\Delta_c\) & - & - & - & - & \negnum{-0.103} & \negnum{-0.037} & \negnum{-0.088} & \negnum{-0.066} & \negnum{-0.058} & \negnum{-0.079} & \negnum{-0.172} & \negnum{-0.045} \\ 
 & \(\Delta_r\) & - & - & - & - & \negnum{-0.092} & \negnum{-0.037} & \negnum{-0.066} & \negnum{-0.055} & \negnum{-0.056} & \negnum{-0.057} & \negnum{-0.151} & \negnum{-0.041} \\ 
\multirow{2}{*}{ENZYMES} & \(\Delta_c\) & \negnum{-0.233} & \negnum{-0.300} & \negnum{-0.433} & \negnum{-0.383} & \negnum{-0.040} & \negnum{-0.022} & \negnum{-0.018} & \negnum{-0.022} & - & - & - & - \\ 
 & \(\Delta_r\) & \negnum{-0.233} & \negnum{-0.383} & \negnum{-0.450} & \negnum{-0.433} & \negnum{-0.036} & \negnum{-0.022} & \negnum{-0.024} & \negnum{-0.018} & - & - & - & - \\ 
\bottomrule
\end{tabular}
}

\resizebox{\columnwidth}{!}{%
\begin{tabular}{llcccc|cccc|cccc}
\toprule
\multicolumn{14}{c}{Classification} \\ 
\cmidrule(lr){1-14}
Dataset & Gap & \multicolumn{4}{c|}{Graph accuracy} & \multicolumn{4}{c|}{Node F1} & \multicolumn{4}{c}{LP AUC} \\ 
\cmidrule(lr){3-6}\cmidrule(lr){7-10}\cmidrule(lr){11-14}
 &  & GAT & GCN & GIN & SAGE & GAT & GCN & GIN & SAGE & GAT & GCN & GIN & SAGE \\ 
\midrule
\multirow{2}{*}{PROTEINS\_full} & \(\Delta_c\) & \negnum{-0.107} & \posnum{+0.009} & \negnum{-0.018} & \negnum{-0.098} & \negnum{-0.142} & \negnum{-0.061} & \negnum{-0.063} & \negnum{-0.336} & - & - & - & - \\ 
 & \(\Delta_r\) & \negnum{-0.071} & \posnum{+0.009} & \negnum{-0.134} & \posnum{+0.018} & \negnum{-0.114} & \negnum{-0.047} & \negnum{-0.061} & \negnum{-0.344} & - & - & - & - \\ 
\multirow{2}{*}{Cora} & \(\Delta_c\) & - & - & - & - & \negnum{-0.026} & \negnum{-0.022} & \posnum{+0.000} & \negnum{-0.015} & \negnum{-0.021} & \negnum{-0.069} & \negnum{-0.018} & \negnum{-0.051} \\ 
 & \(\Delta_r\) & - & - & - & - & \negnum{-0.018} & \negnum{-0.037} & \posnum{+0.004} & \negnum{-0.004} & \negnum{-0.020} & \negnum{-0.086} & \negnum{-0.005} & \negnum{-0.048} \\ 
\multirow{2}{*}{ENZYMES} & \(\Delta_c\) & \negnum{-0.183} & \negnum{-0.183} & \negnum{-0.333} & \negnum{-0.350} & \negnum{-0.015} & \posnum{+0.005} & \negnum{-0.046} & \negnum{-0.083} & - & - & - & - \\ 
 & \(\Delta_r\) & \negnum{-0.167} & \negnum{-0.250} & \negnum{-0.317} & \negnum{-0.383} & \negnum{-0.008} & \posnum{+0.002} & \negnum{-0.049} & \negnum{-0.076} & - & - & - & - \\ 
\bottomrule
\end{tabular}
}
\end{table}
\begin{table}[!htbp]
\centering
\caption{PROTEINS\_full objective-design ablation: graph accuracy / node F1 gaps relative to clean and random baselines.}
\label{tab:appendix-proteins-objective-design}
\renewcommand{\posnum}[1]{\textbf{#1}}
\resizebox{\columnwidth}{!}{%
\begin{tabular}{llcccc|cccc}
\toprule
\multicolumn{10}{c}{TASO} \\
\cmidrule(lr){1-10}
 &  & \multicolumn{4}{c|}{Graph accuracy} & \multicolumn{4}{c}{Node F1} \\
\cmidrule(lr){3-6}\cmidrule(lr){7-10}
Gap & Source $\backslash$ Target & GIN & GAT & GCN & SAGE & GIN & GAT & GCN & SAGE \\
\midrule
\multirow{4}{*}{\(\Delta_c\)} & GIN & \negnum{-0.170} & \negnum{-0.250} & \negnum{-0.179} & \negnum{-0.286} & \negnum{-0.036} & \negnum{-0.144} & \negnum{-0.040} & \negnum{-0.231} \\
 & GAT & \negnum{-0.054} & \negnum{-0.205} & \negnum{-0.170} & \negnum{-0.045} & \negnum{-0.055} & \negnum{-0.139} & \negnum{-0.062} & \negnum{-0.276} \\
 & GCN & \negnum{-0.107} & \negnum{-0.232} & \negnum{-0.161} & \negnum{-0.241} & \negnum{-0.047} & \negnum{-0.158} & \negnum{-0.011} & \negnum{-0.309} \\
 & SAGE & \negnum{-0.232} & \negnum{-0.134} & \posnum{+0.045} & \negnum{-0.250} & \negnum{-0.017} & \negnum{-0.030} & \negnum{-0.020} & \negnum{-0.140} \\
\cmidrule(lr){1-10}
\multirow{4}{*}{\(\Delta_r\)} & GIN & \negnum{-0.143} & \negnum{-0.152} & \negnum{-0.188} & \negnum{-0.304} & \negnum{-0.019} & \negnum{-0.133} & \negnum{-0.030} & \negnum{-0.220} \\
 & GAT & \negnum{-0.045} & \negnum{-0.152} & \negnum{-0.179} & \negnum{-0.063} & \negnum{-0.046} & \negnum{-0.118} & \negnum{-0.055} & \negnum{-0.265} \\
 & GCN & \negnum{-0.205} & \negnum{-0.143} & \negnum{-0.152} & \negnum{-0.232} & \negnum{-0.040} & \negnum{-0.155} & \negnum{-0.004} & \negnum{-0.294} \\
 & SAGE & \negnum{-0.321} & \negnum{-0.161} & \posnum{+0.071} & \negnum{-0.250} & \negnum{-0.008} & \negnum{-0.019} & \negnum{-0.013} & \negnum{-0.123} \\
\bottomrule
\end{tabular}
}

\resizebox{\columnwidth}{!}{%
\begin{tabular}{llcccc|cccc}
\toprule
\multicolumn{10}{c}{Classification} \\
\cmidrule(lr){1-10}
 &  & \multicolumn{4}{c|}{Graph accuracy} & \multicolumn{4}{c}{Node F1} \\
\cmidrule(lr){3-6}\cmidrule(lr){7-10}
Gap & Source $\backslash$ Target & GIN & GAT & GCN & SAGE & GIN & GAT & GCN & SAGE \\
\midrule
\multirow{4}{*}{\(\Delta_c\)} & GIN & \negnum{-0.018} & \negnum{-0.098} & \negnum{-0.071} & \posnum{+0.000} & \negnum{-0.063} & \negnum{-0.046} & \negnum{-0.043} & \negnum{-0.128} \\
 & GAT & \negnum{-0.009} & \negnum{-0.107} & \negnum{-0.125} & \negnum{-0.116} & \negnum{-0.064} & \negnum{-0.142} & \negnum{-0.088} & \negnum{-0.152} \\
 & GCN & \posnum{+0.098} & \negnum{-0.134} & \posnum{+0.009} & \negnum{-0.232} & \negnum{-0.073} & \negnum{-0.089} & \negnum{-0.061} & \negnum{-0.135} \\
 & SAGE & \posnum{+0.009} & \posnum{+0.018} & \negnum{-0.054} & \negnum{-0.098} & \negnum{-0.024} & \negnum{-0.162} & \negnum{-0.013} & \negnum{-0.336} \\
\cmidrule(lr){1-10}
\multirow{4}{*}{\(\Delta_r\)} & GIN & \negnum{-0.134} & \negnum{-0.054} & \negnum{-0.071} & \posnum{+0.027} & \negnum{-0.061} & \negnum{-0.021} & \negnum{-0.029} & \negnum{-0.095} \\
 & GAT & \negnum{-0.134} & \negnum{-0.071} & \negnum{-0.170} & \negnum{-0.134} & \negnum{-0.050} & \negnum{-0.114} & \negnum{-0.081} & \negnum{-0.127} \\
 & GCN & \negnum{-0.063} & \negnum{-0.045} & \posnum{+0.009} & \negnum{-0.214} & \negnum{-0.071} & \negnum{-0.071} & \negnum{-0.047} & \negnum{-0.101} \\
 & SAGE & \negnum{-0.027} & \posnum{+0.009} & \negnum{-0.054} & \posnum{+0.018} & \negnum{-0.018} & \negnum{-0.155} & \negnum{-0.007} & \negnum{-0.344} \\
\bottomrule
\end{tabular}
}
\end{table}
\subsubsection{Discrete attribute flip optimizer}
This ablation evaluates TAP hard-coordinate search for discrete node-attribute perturbations. As a baseline, we use a relaxed categorical procedure~\citep{liu2023graphcloak} that performs softmax PGD, samples hard categories from the resulting probabilities, and enforces the same graph-wise flip budget. As shown in Table~\ref{tab:mutag-discrete-flip-transfer-matrices}
, TAP achieves stronger and more consistently negative degradation gaps than the relaxed PGD baseline.

\begin{table}[!htbp]
\centering
\renewcommand{\posnum}[1]{\textbf{#1}}
\caption{MUTAG discrete-flip optimizer ablation: same-backbone graph-accuracy gaps relative to clean and budget-matched random categorical-flip baselines.}
\label{tab:mutag-discrete-flip-transfer-matrices}
\resizebox{\columnwidth}{!}{
\begin{tabular}{lcccc}
\toprule
\multicolumn{5}{c}{Hard-coordinate atomic flips} \\
\midrule
Gap & GAT & GCN & GIN & SAGE \\ 
\midrule
\(\Delta_c\) & \negnum{\meanstd{-0.233}{0.076}} & \negnum{\meanstd{-0.200}{0.050}} & \negnum{\meanstd{-0.217}{0.058}} & \negnum{\meanstd{-0.250}{0.050}} \\ 
\(\Delta_r\) & \negnum{\meanstd{-0.200}{0.050}} & \negnum{\meanstd{-0.100}{0.050}} & \negnum{\meanstd{-0.167}{0.076}} & \negnum{\meanstd{-0.183}{0.029}} \\ 
\bottomrule
\end{tabular}
}
\par\medskip
\resizebox{\columnwidth}{!}{
\begin{tabular}{lcccc}
\toprule
\multicolumn{5}{c}{Relaxed PGD + projection} \\
\midrule
Gap & GAT & GCN & GIN & SAGE \\ 
\midrule
\(\Delta_c\) & \negnum{\meanstd{-0.167}{0.104}} & \negnum{\meanstd{-0.133}{0.058}} & \negnum{\meanstd{-0.067}{0.029}} & \negnum{\meanstd{-0.183}{0.076}} \\ 
\(\Delta_r\) & \negnum{\meanstd{-0.133}{0.104}} & \posnum{\meanstd{-0.033}{0.058}} & \posnum{\meanstd{-0.017}{0.076}} & \negnum{\meanstd{-0.117}{0.076}} \\ 
\bottomrule
\end{tabular}
}
\end{table}

\subsection{Complexity Analysis}

MUGEN's perturbation optimization scales linearly with the number and size of the input graphs, including their nodes, edges, and sampled link pairs. Detailed complexity analysis and empirical running times are provided in Appendix~\ref{app:complexity}.

%% file: sections/conclusion.tex
\FloatBarrier
\section{Conclusion and Limitations}
\label{sec:conclusion}

Graph-data owners may release a dataset without knowing whether it will be used for node classification, graph classification, or link prediction. We introduce MUGEN to generate one feature-perturbed release that protects the enabled tasks. To construct this release, we use TASO to introduce classwise-separability signals into perturbation optimization, strengthening unlearnability and transfer across GNN backbones, and TAP to tailor updates to attribute type, substantially strengthening unlearnability for categorical attributes. Across five benchmarks, we show that the resulting perturbations degrade clean-test performance relative to clean and matched random-perturbation training, transfer across GNN backbones, and remain effective under the studied self-supervised objectives and training interventions. For link prediction, we consider binary labels and one link-representation aggregation; extending MUGEN to relation prediction in knowledge graphs is a valuable direction.

\section*{Ethical Considerations}

MUGEN is designed to help data owners protect released graph data from
unauthorized exploitation. Our experiments use publicly available benchmark
datasets and do not process private personal data. The method perturbs node
features while preserving graph topology and task annotations. In practical
deployments, perturbations may affect legitimate downstream users; therefore,
MUGEN should be applied by authorized data owners or data stewards, with its
utility evaluated on intended downstream tasks before release.

%% file: sections/appendix_1.tex
\section{Appendix}

\subsection{Computational Complexity and Runtime}
\label{app:complexity}

Consider a batch containing \(G\) graphs, \(N\) nodes, \(|\mathcal E|\) edges, feature dimension \(d\), and \(P\) sampled link pairs. The encoder has \(L\) message-passing layers with hidden width \(h\), giving
\[
\mathcal C_{\mathrm{enc}}
=
\mathcal O\!\left(
Ndh+|\mathcal E|h+(L-1)(Nh^2+|\mathcal E|h)
\right).
\]
Let \(\mathcal C_{\mathrm{head}}\) denote the total cost of the enabled graph-, node-, and link-prediction heads. For task \(t\), let \(M_t\) be the number of task representations, \(r_t\) their dimension, \(C_t\) the number of classes, and \(J_t\) the number of representation sets to which CSD is applied. TASO introduces the additional cost
\[
\mathcal C_{\mathrm{CSD}}
=
\mathcal O\!\left(
\sum_{t\in\mathcal T}
J_t\left[C_tM_t+(M_t+C_t^2)r_t\right]
\right),
\]
where \(M_t\) equals \(G\), \(N\), or \(P\) for graph, node, or link prediction, respectively. Let \(\mathcal{C}_{\mathrm{head}}\) denote the total cost of the enabled prediction heads. If \(\tau\) encoder evaluations are required, one perturbation-loss evaluation costs
\(\mathcal C_{\mathrm{pert}} = \mathcal O\!\left(\tau\mathcal C_{\mathrm{enc}} + \mathcal C_{\mathrm{head}} + \mathcal C_{\mathrm{CSD}}\right)\).

Let \(S\) be the continuous PGD steps per update, \(Q\) the categorical coordinate steps, \(K\) the exact-candidate limit, and \(C_{\mathrm{cat}}\) the number of attribute categories. One continuous perturbation update costs
\[
\mathcal O\!\left(
S(\mathcal C_{\mathrm{pert}}+Nd)
\right),
\]
whereas one categorical perturbation update costs
\[
\mathcal O\!\left(
Q\bigl((K+2)\mathcal C_{\mathrm{pert}}+NC_{\mathrm{cat}}\bigr)
\right).
\]
Thus, TAP’s optimization is linearly scalable with input size, including the number of graphs and each graph’s nodes, edges, and sampled link pairs.

Table~\ref{tab:creation-runtime} reports the one-off empirical perturbation-creation time for each source backend. All datasets except MUTAG were run on a single NVIDIA RTX 3090 GPU; MUTAG was run on a CPU.

\begin{table}[htbp]
\centering
\caption{Empirical perturbation-creation time (s) for each source backend.}
\label{tab:creation-runtime}
\resizebox{\columnwidth}{!}{%
\begin{tabular}{lrrrrr}
\toprule
Dataset & Opt. epochs & GAT & GCN & GIN & GraphSAGE \\
\midrule
PROTEINS\_full & 280 & 511.4 & 512.1 & 463.3 & 443.9 \\
ENZYMES & 280 & 700.9 & 535.2 & 479.7 & 509.7 \\
Cora & 200 & 130.7 & 95.8 & 98.9 & 96.2 \\
PubMed & 200 & 172.7 & 141.0 & 168.8 & 153.2 \\
MUTAG & 180 & 31.2 & 25.8 & 38.5 & 15.3 \\
\bottomrule
\end{tabular}
}
\end{table}

%% file: sections/appendix.tex
\raggedbottom
\subsection{Complete Robustness Results}
\label{app:complete-robustness-results}

\begin{figure*}[t]
\centering
\includegraphics[width=0.70\textwidth]{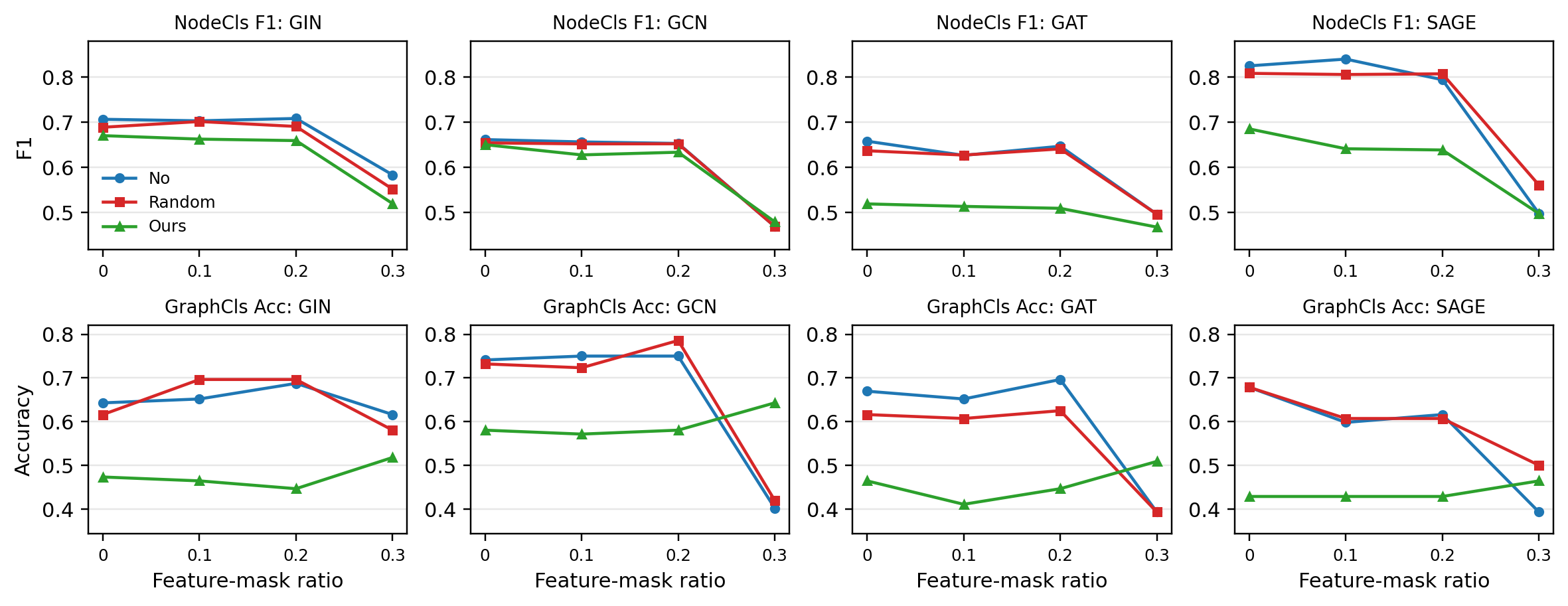}
\caption{Complete feature-masking robustness results on PROTEINS\_full. Panels show node F1 and graph accuracy for clean data (No), Gaussian random perturbations (Random), and perturbations generated by our framework (Ours).}
\Description{A complete panel of PROTEINS_full robustness curves under feature masking compares node F1 and graph accuracy for clean data, Gaussian random perturbations, and MUGEN perturbations.}
\label{fig:app-full-fmask-proteins}
\end{figure*}

\begin{figure*}[t]
\centering
\includegraphics[width=0.70\textwidth]{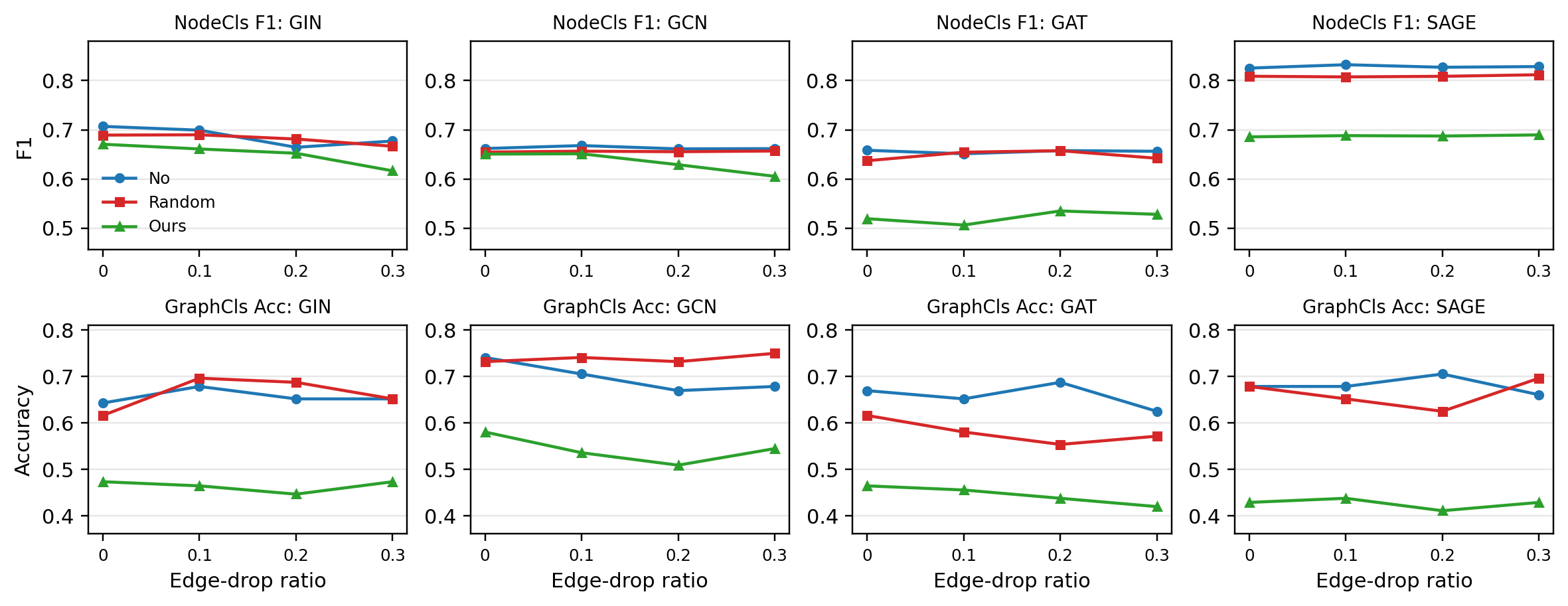}
\caption{Complete edge-dropping robustness results on PROTEINS\_full. Panels show node F1 and graph accuracy for clean data (No), Gaussian random perturbations (Random), and perturbations generated by our framework (Ours).}
\Description{A complete panel of PROTEINS_full robustness curves under edge dropping compares node F1 and graph accuracy for clean data, Gaussian random perturbations, and MUGEN perturbations.}
\label{fig:app-full-edrop-proteins}
\end{figure*}

\begin{figure*}[t]
\centering
\includegraphics[width=0.70\textwidth]{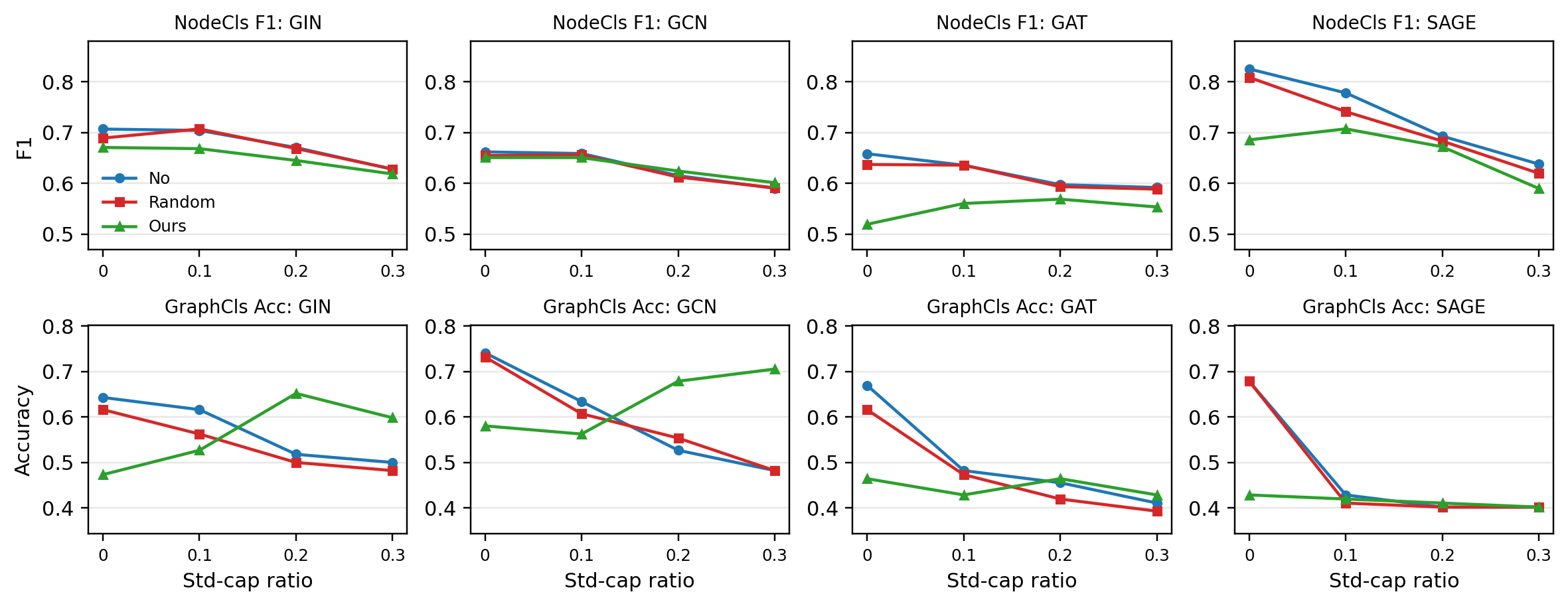}
\caption{Complete adversarial-training robustness results on PROTEINS\_full. Panels show node F1 and graph accuracy for clean data (No), Gaussian random perturbations (Random), and perturbations generated by our framework (Ours).}
\Description{A complete panel of PROTEINS_full robustness curves under adversarial training compares node F1 and graph accuracy for clean data, Gaussian random perturbations, and MUGEN perturbations.}
\label{fig:app-full-advtrain-proteins}
\end{figure*}

We provide the complete PROTEINS\_full robustness results, including node F1 and graph accuracy, for feature masking (Figure~\ref{fig:app-full-fmask-proteins}), edge dropping (Figure~\ref{fig:app-full-edrop-proteins}), and adversarial training (Figure~\ref{fig:app-full-advtrain-proteins}).

\subsection{Other Complete Results for Objective Design for Transferability}
\label{app:objective-design-ablation}
\noindent Tables~\ref{tab:appendix-cora-objective-design} and~\ref{tab:enzymes-objective-design} provide the complete four-source-by-four-target transfer matrices for Cora and ENZYMES, respectively. They complement the diagonal summary in Table~\ref{tab:objective-design-diagonal}; lower accuracy indicates stronger unlearnability.
\vspace{0.5\baselineskip}
\begin{table*}[t]
\centering
\caption{Cora objective-design ablation: node F1 / LP AUC / LP AP gaps relative to clean and random baselines.}
\label{tab:appendix-cora-objective-design}
\renewcommand{\posnum}[1]{\textbf{#1}}
\resizebox{\textwidth}{!}{%
\begin{tabular}{@{}c@{\hspace{0.6em}}c@{}}
\begin{tabular}{llcccc|cccc|cccc}
\toprule
\multicolumn{14}{c}{TASO} \\
\cmidrule(lr){1-14}
 &  & \multicolumn{4}{c|}{Node F1} & \multicolumn{4}{c|}{LP AUC} & \multicolumn{4}{c}{LP AP} \\
\cmidrule(lr){3-6}\cmidrule(lr){7-10}\cmidrule(lr){11-14}
Gap & Source $\backslash$ Target & GAT & GCN & GIN & SAGE & GAT & GCN & GIN & SAGE & GAT & GCN & GIN & SAGE \\
\midrule
\multirow{4}{*}{\(\Delta_c\)} & GAT & \negnum{-0.103} & \negnum{-0.088} & \negnum{-0.052} & \negnum{-0.044} & \negnum{-0.058} & \negnum{-0.139} & \negnum{-0.095} & \negnum{-0.054} & \negnum{-0.053} & \negnum{-0.153} & \negnum{-0.088} & \negnum{-0.055} \\
 & GCN & \negnum{-0.088} & \negnum{-0.037} & \negnum{-0.066} & \negnum{-0.085} & \negnum{-0.118} & \negnum{-0.079} & \negnum{-0.057} & \negnum{-0.050} & \negnum{-0.087} & \negnum{-0.070} & \negnum{-0.058} & \negnum{-0.053} \\
 & GIN & \negnum{-0.029} & \negnum{-0.052} & \negnum{-0.088} & \negnum{-0.044} & \negnum{-0.069} & \negnum{-0.122} & \negnum{-0.172} & \negnum{-0.032} & \negnum{-0.045} & \negnum{-0.112} & \negnum{-0.159} & \negnum{-0.040} \\
 & SAGE & \negnum{-0.074} & \negnum{-0.018} & \negnum{-0.070} & \negnum{-0.066} & \negnum{-0.018} & \negnum{-0.031} & \negnum{-0.020} & \negnum{-0.045} & \negnum{-0.029} & \negnum{-0.057} & \negnum{-0.036} & \negnum{-0.064} \\
\cmidrule(lr){1-14}
\multirow{4}{*}{\(\Delta_r\)} & GAT & \negnum{-0.092} & \negnum{-0.103} & \negnum{-0.029} & \negnum{-0.033} & \negnum{-0.056} & \negnum{-0.141} & \negnum{-0.081} & \negnum{-0.051} & \negnum{-0.049} & \negnum{-0.158} & \negnum{-0.073} & \negnum{-0.053} \\
 & GCN & \negnum{-0.077} & \negnum{-0.037} & \negnum{-0.044} & \negnum{-0.088} & \negnum{-0.116} & \negnum{-0.057} & \negnum{-0.049} & \negnum{-0.047} & \negnum{-0.085} & \negnum{-0.058} & \negnum{-0.052} & \negnum{-0.051} \\
 & GIN & \negnum{-0.018} & \negnum{-0.074} & \negnum{-0.066} & \negnum{-0.048} & \negnum{-0.067} & \negnum{-0.128} & \negnum{-0.151} & \negnum{-0.029} & \negnum{-0.042} & \negnum{-0.119} & \negnum{-0.133} & \negnum{-0.037} \\
 & SAGE & \negnum{-0.059} & \negnum{-0.018} & \negnum{-0.044} & \negnum{-0.055} & \negnum{-0.012} & \negnum{-0.039} & \negnum{-0.025} & \negnum{-0.041} & \negnum{-0.024} & \negnum{-0.067} & \negnum{-0.042} & \negnum{-0.061} \\
\bottomrule
\end{tabular}
&
\begin{tabular}{llcccc|cccc|cccc}
\toprule
\multicolumn{14}{c}{Classification} \\
\cmidrule(lr){1-14}
 &  & \multicolumn{4}{c|}{Node F1} & \multicolumn{4}{c|}{LP AUC} & \multicolumn{4}{c}{LP AP} \\
\cmidrule(lr){3-6}\cmidrule(lr){7-10}\cmidrule(lr){11-14}
Gap & Source $\backslash$ Target & GAT & GCN & GIN & SAGE & GAT & GCN & GIN & SAGE & GAT & GCN & GIN & SAGE \\
\midrule
\multirow{4}{*}{\(\Delta_c\)} & GAT & \negnum{-0.026} & \negnum{-0.004} & \negnum{-0.011} & \negnum{-0.015} & \negnum{-0.021} & \negnum{-0.062} & \negnum{-0.174} & \negnum{-0.020} & \negnum{-0.031} & \negnum{-0.032} & \negnum{-0.186} & \negnum{-0.022} \\
 & GCN & \negnum{-0.018} & \negnum{-0.022} & \negnum{-0.004} & \negnum{-0.022} & \negnum{-0.020} & \negnum{-0.069} & \negnum{-0.078} & \negnum{-0.023} & \negnum{-0.030} & \negnum{-0.059} & \negnum{-0.079} & \negnum{-0.026} \\
 & GIN & \negnum{-0.011} & \posnum{+0.018} & \posnum{+0.000} & \negnum{-0.011} & \negnum{-0.019} & \posnum{+0.031} & \negnum{-0.018} & \negnum{-0.008} & \negnum{-0.028} & \posnum{+0.024} & \negnum{-0.046} & \negnum{-0.017} \\
 & SAGE & \negnum{-0.018} & \negnum{-0.007} & \negnum{-0.004} & \negnum{-0.015} & \negnum{-0.026} & \negnum{-0.013} & \negnum{-0.050} & \negnum{-0.051} & \negnum{-0.029} & \negnum{-0.025} & \negnum{-0.051} & \negnum{-0.059} \\
\cmidrule(lr){1-14}
\multirow{4}{*}{\(\Delta_r\)} & GAT & \negnum{-0.018} & \negnum{-0.004} & \negnum{-0.011} & \negnum{-0.018} & \negnum{-0.020} & \negnum{-0.068} & \negnum{-0.174} & \negnum{-0.017} & \negnum{-0.029} & \negnum{-0.043} & \negnum{-0.188} & \negnum{-0.020} \\
 & GCN & \negnum{-0.011} & \negnum{-0.037} & \negnum{-0.004} & \negnum{-0.026} & \negnum{-0.019} & \negnum{-0.086} & \negnum{-0.052} & \negnum{-0.020} & \negnum{-0.027} & \negnum{-0.075} & \negnum{-0.048} & \negnum{-0.024} \\
 & GIN & \posnum{+0.000} & \posnum{+0.018} & \posnum{+0.004} & \negnum{-0.015} & \negnum{-0.019} & \posnum{+0.017} & \negnum{-0.005} & \negnum{-0.005} & \negnum{-0.027} & \posnum{+0.009} & \negnum{-0.020} & \negnum{-0.015} \\
 & SAGE & \negnum{-0.011} & \negnum{-0.007} & \posnum{+0.007} & \negnum{-0.004} & \negnum{-0.025} & \negnum{-0.021} & \negnum{-0.032} & \negnum{-0.048} & \negnum{-0.026} & \negnum{-0.041} & \negnum{-0.036} & \negnum{-0.057} \\
\bottomrule
\end{tabular}
\end{tabular}
}
\end{table*}

\begin{table*}[t]
\centering
\caption{ENZYMES objective-design ablation: graph accuracy / node F1 gaps relative to clean and random baselines.}
\label{tab:enzymes-objective-design}
\renewcommand{\posnum}[1]{\textbf{#1}}
\resizebox{\textwidth}{!}{%
\begin{tabular}{@{}c@{\hspace{0.6em}}c@{}}
\begin{tabular}{llcccc|cccc}
\toprule
\multicolumn{10}{c}{TASO} \\
\cmidrule(lr){1-10}
 &  & \multicolumn{4}{c|}{Graph accuracy} & \multicolumn{4}{c}{Node F1} \\
\cmidrule(lr){3-6}\cmidrule(lr){7-10}
Gap & Source $\backslash$ Target & GIN & GAT & GCN & SAGE & GIN & GAT & GCN & SAGE \\
\midrule
\multirow{4}{*}{\(\Delta_c\)} & GIN & \negnum{-0.433} & \negnum{-0.250} & \negnum{-0.233} & \negnum{-0.517} & \negnum{-0.018} & \negnum{-0.035} & \negnum{-0.031} & \negnum{-0.012} \\
 & GAT & \negnum{-0.333} & \negnum{-0.233} & \negnum{-0.233} & \negnum{-0.450} & \negnum{-0.032} & \negnum{-0.040} & \negnum{-0.026} & \negnum{-0.013} \\
 & GCN & \negnum{-0.350} & \negnum{-0.067} & \negnum{-0.300} & \negnum{-0.500} & \negnum{-0.034} & \negnum{-0.040} & \negnum{-0.022} & \negnum{-0.017} \\
 & SAGE & \negnum{-0.450} & \negnum{-0.117} & \negnum{-0.283} & \negnum{-0.383} & \negnum{-0.023} & \negnum{-0.011} & \posnum{+0.000} & \negnum{-0.022} \\
\cmidrule(lr){1-10}
\multirow{4}{*}{\(\Delta_r\)} & GIN & \negnum{-0.450} & \negnum{-0.233} & \negnum{-0.217} & \negnum{-0.550} & \negnum{-0.024} & \negnum{-0.032} & \negnum{-0.032} & \negnum{-0.008} \\
 & GAT & \negnum{-0.383} & \negnum{-0.233} & \negnum{-0.300} & \negnum{-0.483} & \negnum{-0.038} & \negnum{-0.036} & \negnum{-0.027} & \negnum{-0.009} \\
 & GCN & \negnum{-0.367} & \negnum{-0.100} & \negnum{-0.383} & \negnum{-0.533} & \negnum{-0.041} & \negnum{-0.038} & \negnum{-0.022} & \negnum{-0.013} \\
 & SAGE & \negnum{-0.467} & \negnum{-0.133} & \negnum{-0.267} & \negnum{-0.433} & \negnum{-0.027} & \negnum{-0.011} & \negnum{-0.002} & \negnum{-0.018} \\
\bottomrule
\end{tabular}
&
\begin{tabular}{llcccc|cccc}
\toprule
\multicolumn{10}{c}{Classification} \\
\cmidrule(lr){1-10}
 &  & \multicolumn{4}{c|}{Graph accuracy} & \multicolumn{4}{c}{Node F1} \\
\cmidrule(lr){3-6}\cmidrule(lr){7-10}
Gap & Source $\backslash$ Target & GIN & GAT & GCN & SAGE & GIN & GAT & GCN & SAGE \\
\midrule
\multirow{4}{*}{\(\Delta_c\)} & GIN & \negnum{-0.333} & \negnum{-0.150} & \negnum{-0.200} & \negnum{-0.367} & \negnum{-0.046} & \posnum{+0.010} & \negnum{-0.006} & \negnum{-0.029} \\
 & GAT & \negnum{-0.367} & \negnum{-0.183} & \negnum{-0.183} & \negnum{-0.417} & \negnum{-0.016} & \negnum{-0.015} & \posnum{+0.004} & \negnum{-0.023} \\
 & GCN & \negnum{-0.333} & \negnum{-0.167} & \negnum{-0.183} & \negnum{-0.283} & \negnum{-0.017} & \negnum{-0.008} & \posnum{+0.005} & \negnum{-0.027} \\
 & SAGE & \negnum{-0.317} & \negnum{-0.133} & \negnum{-0.100} & \negnum{-0.350} & \negnum{-0.002} & \negnum{-0.046} & \negnum{-0.005} & \negnum{-0.083} \\
\cmidrule(lr){1-10}
\multirow{4}{*}{\(\Delta_r\)} & GIN & \negnum{-0.317} & \negnum{-0.133} & \negnum{-0.300} & \negnum{-0.367} & \negnum{-0.049} & \posnum{+0.021} & \negnum{-0.011} & \negnum{-0.022} \\
 & GAT & \negnum{-0.400} & \negnum{-0.167} & \negnum{-0.233} & \negnum{-0.433} & \negnum{-0.023} & \negnum{-0.008} & \posnum{+0.006} & \negnum{-0.016} \\
 & GCN & \negnum{-0.383} & \negnum{-0.133} & \negnum{-0.250} & \negnum{-0.283} & \negnum{-0.022} & \posnum{+0.001} & \posnum{+0.002} & \negnum{-0.020} \\
 & SAGE & \negnum{-0.317} & \negnum{-0.100} & \negnum{-0.133} & \negnum{-0.383} & \negnum{-0.009} & \negnum{-0.037} & \negnum{-0.007} & \negnum{-0.076} \\
\bottomrule
\end{tabular}
\end{tabular}
}
\end{table*}

\subsection{Continuous Node-Feature Distribution Diagnostics}

\begin{figure*}[t]
\centering
\includegraphics[width=\textwidth]{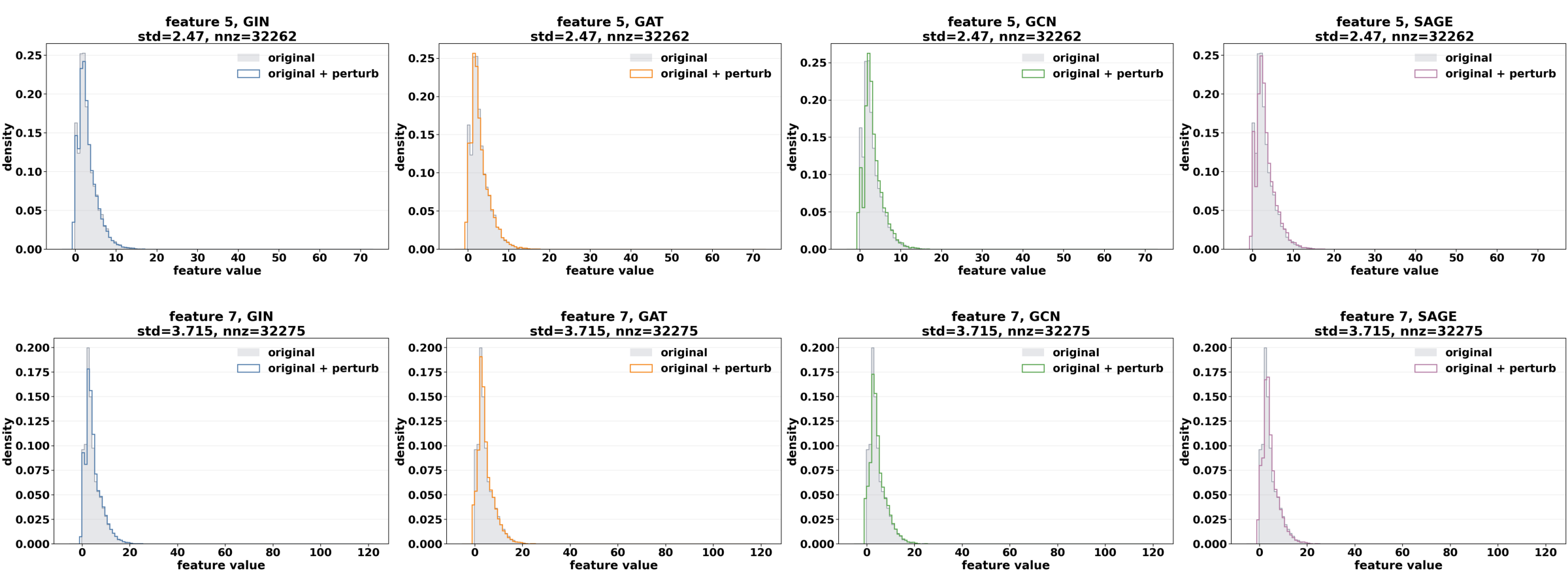}
\caption{Original versus perturbed feature-value histograms for PROTEINS\_full, shown for representative features.}
\Description{A grid of histograms compares original and perturbed values for representative PROTEINS_full node features, showing the distributional changes introduced by the perturbation.}
\label{fig:exp170-feature-histograms}
\end{figure*}

\begin{figure*}[t]
\centering
\includegraphics[width=\textwidth]{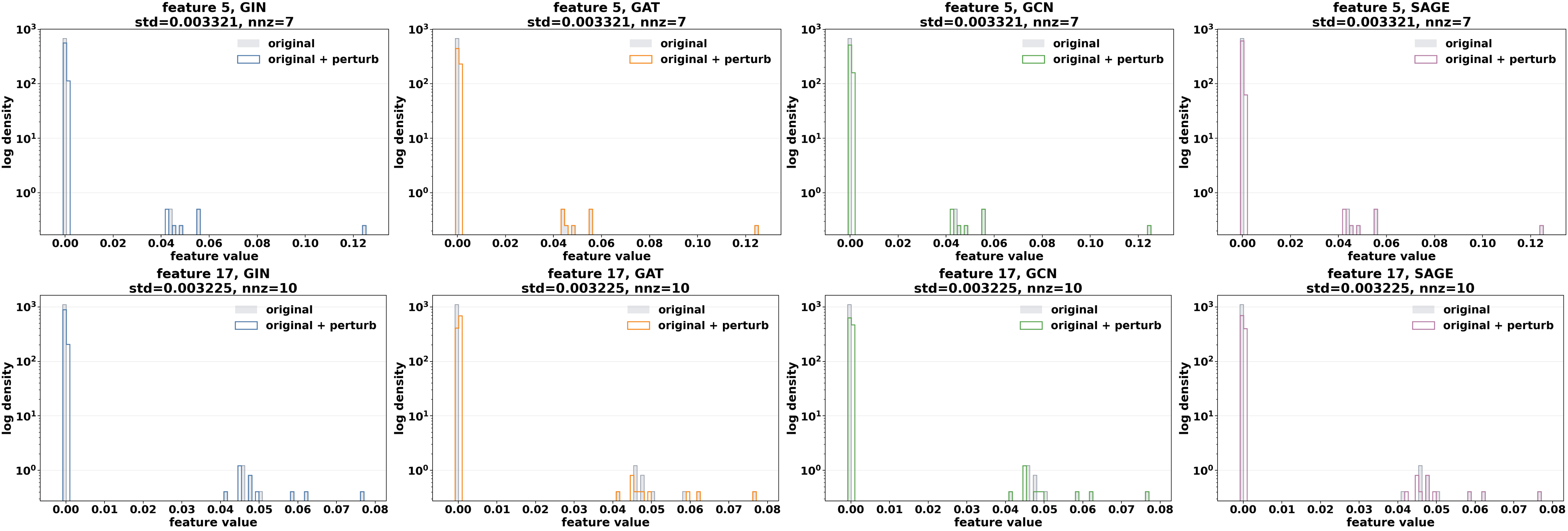}
\caption{Original versus perturbed feature-value histograms for Cora, shown for representative features. The logarithmic y-axis improves visibility of low-frequency nonzero values because the original feature values are highly concentrated near zero.}
\Description{A grid of logarithmic-scale histograms compares original and perturbed values for representative Cora features. The display makes rare nonzero values visible despite the strong concentration of original values near zero.}
\label{fig:exp183-cora-feature-run0}
\end{figure*}

\begin{figure*}[t]
\centering
\includegraphics[width=\textwidth]{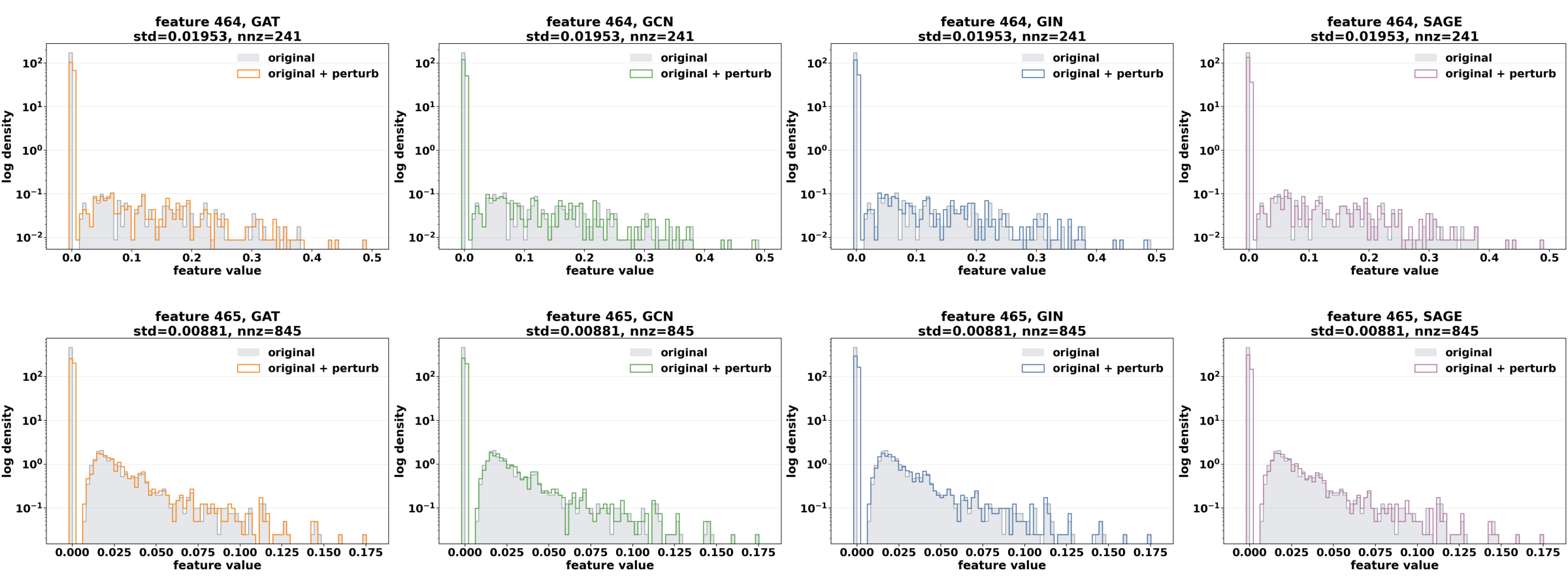}
\caption{Original versus perturbed feature-value histograms for PubMed, shown for representative features.}
\Description{A grid of histograms compares original and perturbed values for representative PubMed node features, showing how the optimized perturbation changes their distributions.}
\label{fig:exp184-pubmed-feature-diagnostics}
\end{figure*}

\begin{figure*}[t]
\centering
\includegraphics[width=\textwidth]{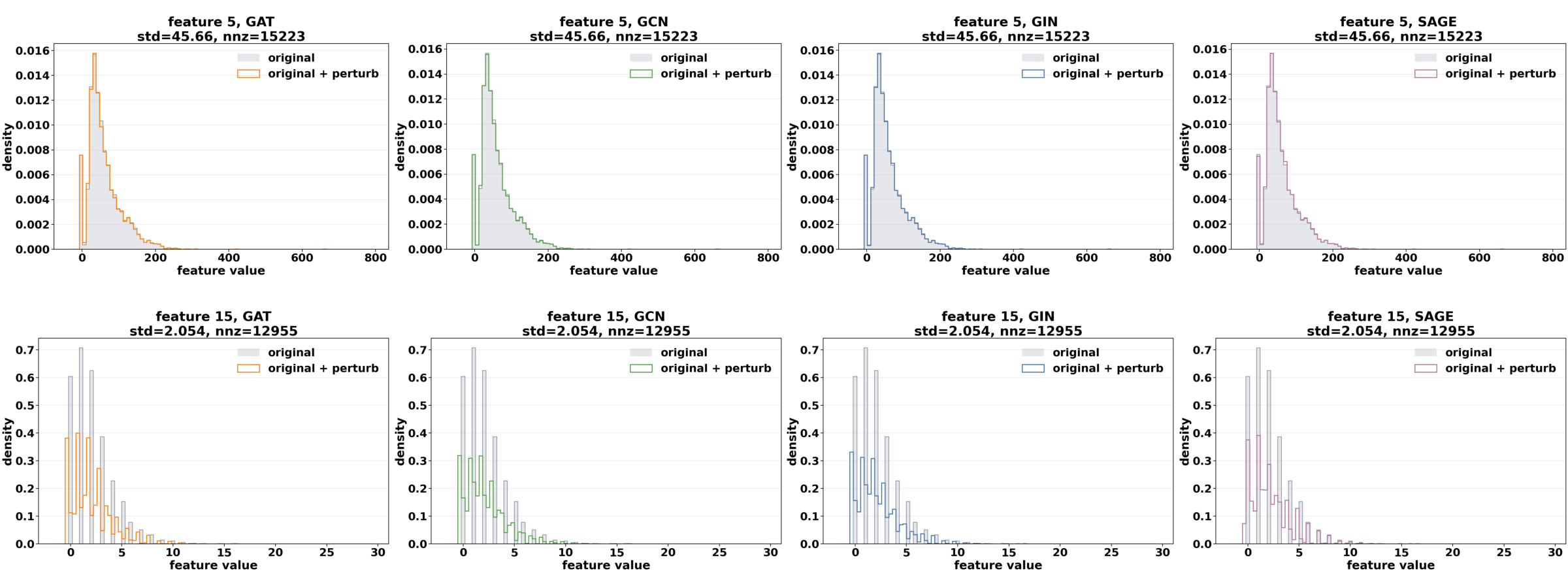}
\caption{Original versus perturbed feature-value histograms for ENZYMES, shown for representative features.}
\Description{A grid of histograms compares original and perturbed values for representative ENZYMES node features, showing the distributional effects of the perturbation.}
\label{fig:exp171-enzymes-feature-diagnostics}
\end{figure*}

We compare original and perturbed continuous node-feature values across representative feature coordinates and datasets: PROTEINS\_full (Figure~\ref{fig:exp170-feature-histograms}), Cora (Figure~\ref{fig:exp183-cora-feature-run0}), PubMed (Figure~\ref{fig:exp184-pubmed-feature-diagnostics}), and ENZYMES (Figure~\ref{fig:exp171-enzymes-feature-diagnostics}).

%% file: references.bib
@inproceedings{ren2022transferable,
  title={Transferable Unlearnable Examples},
  author={Ren, Jie and Xu, Han and Wan, Yuxuan and Ma, Xingjun and Sun, Lichao and Tang, Jiliang},
  booktitle={International Conference on Learning Representations},
  year={2023},
  url={https://arxiv.org/abs/2210.10114}
}

@article{zhu2023spottarget,
  title={Pitfalls in Link Prediction with Graph Neural Networks: Understanding the Impact of Target-link Inclusion and Better Practices},
  author={Zhu, Jing and Zhou, Yuhang and Ioannidis, Vassilis N. and Qian, Shengyi and Ai, Wei and Song, Xiang and Koutra, Danai},
  journal={arXiv preprint arXiv:2306.00899},
  year={2023},
  url={https://arxiv.org/abs/2306.00899}
}

@inproceedings{yu2025mtlue,
  title={{MTL-UE}: Learning to Learn Nothing for Multi-Task Learning},
  author={Yu, Yi and Xia, Song and Yang, Siyuan and Kong, Chenqi and Yang, Wenhan and Lu, Shijian and Tan, Yap-Peng and Kot, Alex},
  booktitle={Proceedings of the 42nd International Conference on Machine Learning},
  series={Proceedings of Machine Learning Research},
  volume={267},
  pages={73286--73303},
  year={2025},
  publisher={PMLR},
  url={https://proceedings.mlr.press/v267/yu25r.html}
}

@inproceedings{madry2018towards,
  title={Towards Deep Learning Models Resistant to Adversarial Attacks},
  author={Madry, Aleksander and Makelov, Aleksandar and Schmidt, Ludwig and Tsipras, Dimitris and Vladu, Adrian},
  booktitle={International Conference on Learning Representations},
  year={2018},
  url={https://openreview.net/forum?id=rJzIBfZAb}
}

@article{li2025surveyunlearnable,
  title={A Survey on Unlearnable Data},
  author={Li, Jiahao and Chen, Yiqiang and Xing, Yunbing and Gu, Yang and Lan, Xiangyuan},
  journal={arXiv preprint arXiv:2503.23536},
  year={2025},
  url={https://arxiv.org/abs/2503.23536}
}

@article{liu2023graphcloak,
  title={GraphCloak: Safeguarding Task-specific Knowledge within Graph-structured Data from Unauthorized Exploitation},
  author={Liu, Yixin and Fan, Chenrui and Chen, Xun and Zhou, Pan and Sun, Lichao},
  journal={arXiv preprint arXiv:2310.07100},
  year={2023},
  url={https://arxiv.org/abs/2310.07100}
}

@inproceedings{huang2021unlearnable,
  title={Unlearnable Examples: Making Personal Data Unexploitable},
  author={Huang, Hanxun and Ma, Xingjun and Erfani, Sarah Monazam and Bailey, James and Wang, Yisen},
  booktitle={International Conference on Learning Representations},
  year={2021},
  url={https://openreview.net/forum?id=iAmZUo0DxC0}
}

@inproceedings{fu2022robust,
  title={Robust Unlearnable Examples: Protecting Data Privacy Against Adversarial Learning},
  author={Fu, Shaopeng and He, Fengxiang and Liu, Yang and Shen, Li and Tao, Dacheng},
  booktitle={International Conference on Learning Representations},
  year={2022},
  url={https://openreview.net/forum?id=baUQQPwQiAg}
}

@inproceedings{liu2024stable,
  title={Stable Unlearnable Example: Enhancing the Robustness of Unlearnable Examples via Stable Error-Minimizing Noise},
  author={Liu, Yixin and Xu, Kaidi and Chen, Xun and Sun, Lichao},
  booktitle={Proceedings of the AAAI Conference on Artificial Intelligence},
  volume={38},
  pages={3783--3791},
  year={2024},
  doi={10.1609/aaai.v38i4.28169}
}

@inproceedings{wang2025provably,
  title={Provably Unlearnable Data Examples},
  author={Wang, Derui and Xue, Minhui and Li, Bo and Camtepe, Seyit and Zhu, Liming},
  booktitle={Network and Distributed System Security Symposium},
  year={2025},
  url={https://www.ndss-symposium.org/ndss-paper/provably-unlearnable-data-examples/}
}

@inproceedings{chen2024oneforall,
  title={One for All: A Universal Generator for Concept Unlearnability via Multi-Modal Alignment},
  author={Chen, Chaochao and Zhang, Jiaming and Li, Yuyuan and Han, Zhongxuan},
  booktitle={Proceedings of the 41st International Conference on Machine Learning},
  series={Proceedings of Machine Learning Research},
  volume={235},
  pages={7700--7711},
  year={2024},
  publisher={PMLR},
  url={https://proceedings.mlr.press/v235/chen24bc.html}
}

@inproceedings{li2025versatile,
  title={Versatile Transferable Unlearnable Example Generator},
  author={Li, Zhihao and Cai, Jiale and Xu, Gezheng and Zheng, Hao and Li, Qiuyue and Zhou, Fan and Yang, Shichun and Ling, Charles and Wang, Boyu},
  booktitle={Advances in Neural Information Processing Systems},
  volume={38},
  year={2025},
  url={https://proceedings.neurips.cc/paper_files/paper/2025/hash/195c57a5462d917276c7d5ec76e0245d-Abstract-Conference.html}
}

@article{gong2026armor,
  title={{ARMOR}: Shielding Unlearnable Examples against Data Augmentation},
  author={Gong, Xueluan and Wang, Yuji and Chen, Yanjiao and Dong, Haocheng and Li, Yiming and Sun, Mengyuan and Li, Shuaike and Wang, Qian and Chen, Chen},
  journal={IEEE Transactions on Pattern Analysis and Machine Intelligence},
  year={2026},
  doi={10.1109/TPAMI.2026.3652456}
}

@inproceedings{li2026priors,
  title={When Priors Backfire: On the Vulnerability of Unlearnable Examples to Pretraining},
  author={Li, Zhihao and Xu, Gezheng and Cai, Jiale and Fang, Ruiyi and Wu, Di and Lao, Qicheng and Ling, Charles and Wang, Boyu},
  booktitle={International Conference on Learning Representations},
  year={2026},
  url={https://openreview.net/forum?id=MQayd2Nmth}
}

@inproceedings{cai2026fuse,
  title={{FUSE}: Full-spectrum Unlearnable Examples via Spectral Equalization},
  author={Cai, Jiale and Xu, Gezheng and Li, Zhihao and Fang, Ruiyi and Pu, Ruizhi and Wu, Di and Lao, Qicheng and Ling, Charles and Wang, Boyu},
  booktitle={Proceedings of the 43rd International Conference on Machine Learning},
  series={Proceedings of Machine Learning Research},
  volume={306},
  year={2026},
  url={https://arxiv.org/abs/2606.26719}
}

@inproceedings{li2023maketext,
  title={Make Text Unlearnable: Exploiting Effective Patterns to Protect Personal Data},
  author={Li, Xinzhe and Liu, Ming},
  booktitle={Proceedings of the 3rd Workshop on Trustworthy Natural Language Processing},
  pages={249--259},
  year={2023},
  publisher={Association for Computational Linguistics},
  doi={10.18653/v1/2023.trustnlp-1.22}
}

@inproceedings{jiang2024timeseries,
  title={Unlearnable Examples for Time Series},
  author={Jiang, Yujing and Ma, Xingjun and Erfani, Sarah Monazam and Bailey, James},
  booktitle={Advances in Knowledge Discovery and Data Mining},
  series={Lecture Notes in Computer Science},
  volume={14650},
  pages={213--225},
  year={2024},
  publisher={Springer},
  doi={10.1007/978-981-97-2266-2_17}
}

@inproceedings{wang2024pointclouds,
  title={Unlearnable 3D Point Clouds: Class-wise Transformation Is All You Need},
  author={Wang, Xianlong and Li, Minghui and Liu, Wei and Zhang, Hangtao and Hu, Shengshan and Zhang, Yechao and Zhou, Ziqi and Jin, Hai},
  booktitle={Advances in Neural Information Processing Systems},
  volume={37},
  year={2024},
  url={https://proceedings.neurips.cc/paper_files/paper/2024/hash/b3d868b4b5b61b35a849ba6e7a1d4449-Abstract-Conference.html}
}

@inproceedings{zhang2026rendering,
  title={Rendering Data Unlearnable by Exploiting {LLM} Alignment Mechanisms},
  author={Zhang, Ruihan and Sun, Jun},
  booktitle={Proceedings of the 64th Annual Meeting of the Association for Computational Linguistics (Volume 1: Long Papers)},
  pages={40587--40598},
  year={2026},
  publisher={Association for Computational Linguistics},
  doi={10.18653/v1/2026.acl-long.1885}
}

@article{zhao2026mmguard,
  title={To See Is Not to Learn: Protecting Multimodal Data from Unauthorized Fine-Tuning of Large Vision-Language Model},
  author={Zhao, Chengshuai and Tan, Zhen and Li, Dawei and Yu, Zhiyuan and Liu, Huan},
  journal={arXiv preprint arXiv:2605.14291},
  year={2026},
  url={https://arxiv.org/abs/2605.14291}
}

@inproceedings{peng2022learnabilitylock,
  title={Learnability Lock: Authorized Learnability Control Through Adversarial Invertible Transformations},
  author={Peng, Weiqi and Chen, Jinghui},
  booktitle={International Conference on Learning Representations},
  year={2022},
  url={https://openreview.net/forum?id=6VpeS27viTq}
}

@inproceedings{sadasivan2023cuda,
  title={{CUDA}: Convolution-Based Unlearnable Datasets},
  author={Sadasivan, Vinu Sankar and Soltanolkotabi, Mahdi and Feizi, Soheil},
  booktitle={Proceedings of the IEEE/CVF Conference on Computer Vision and Pattern Recognition},
  pages={3862--3871},
  year={2023},
  doi={10.1109/CVPR52729.2023.00376}
}

@inproceedings{wang2024efficient,
  title={Efficient Availability Attacks against Supervised and Contrastive Learning Simultaneously},
  author={Wang, Yihan and Zhu, Yifan and Gao, Xiao-Shan},
  booktitle={Advances in Neural Information Processing Systems},
  volume={37},
  year={2024},
  url={https://proceedings.neurips.cc/paper_files/paper/2024/file/85826ad1eb4602a2962b7cdbe129b341-Paper-Conference.pdf}
}

@inproceedings{liu2023unlearnablegraph,
  title={Unlearnable Graph: Protecting Graphs from Unauthorized Exploitation},
  author={Liu, Yixin and Fan, Chenrui and Zhou, Pan and Sun, Lichao},
  booktitle={Network and Distributed System Security Symposium Poster},
  year={2023},
  url={https://arxiv.org/abs/2303.02568}
}

@inproceedings{ebrahimi-etal-2018-hotflip,
  title={{H}ot{F}lip: White-Box Adversarial Examples for Text Classification},
  author={Ebrahimi, Javid and Rao, Anyi and Lowd, Daniel and Dou, Dejing},
  booktitle={Proceedings of the 56th Annual Meeting of the Association for Computational Linguistics (Volume 2: Short Papers)},
  pages={31--36},
  year={2018},
  address={Melbourne, Australia},
  publisher={Association for Computational Linguistics},
  doi={10.18653/v1/P18-2006},
  url={https://aclanthology.org/P18-2006/}
}

@inproceedings{sypetkowski2024scalability,
  title={On the Scalability of {GNNs} for Molecular Graphs},
  author={Sypetkowski, Maciej and Wenkel, Frederik and Poursafaei, Farimah and Dickson, Nia and Suri, Karush and Fradkin, Philip and Beaini, Dominique},
  booktitle={Advances in Neural Information Processing Systems},
  volume={37},
  year={2024},
  url={https://proceedings.neurips.cc/paper_files/paper/2024/hash/2345275663a15ee92a06bc957be54a2c-Abstract-Conference.html}
}

@inproceedings{wang2024gft,
  title={{GFT}: Graph Foundation Model with Transferable Tree Vocabulary},
  author={Wang, Zehong and Zhang, Zheyuan and Chawla, Nitesh V. and Zhang, Chuxu and Ye, Yanfang},
  booktitle={Advances in Neural Information Processing Systems},
  volume={37},
  year={2024},
  url={https://proceedings.neurips.cc/paper_files/paper/2024/hash/c23ccf9eedf87e4380e92b75b24955bb-Abstract-Conference.html}
}

@inproceedings{wang2025tasktrees,
  title={Towards Graph Foundation Models: Learning Generalities Across Graphs via Task-Trees},
  author={Wang, Zehong and Zhang, Zheyuan and Ma, Tianyi and Chawla, Nitesh V. and Zhang, Chuxu and Ye, Yanfang},
  booktitle={Proceedings of the 42nd International Conference on Machine Learning},
  series={Proceedings of Machine Learning Research},
  volume={267},
  pages={65518--65555},
  year={2025},
  publisher={PMLR},
  url={https://proceedings.mlr.press/v267/wang25eq.html}
}

@inproceedings{dehghan2024ewek,
  title={{EWEK}-{QA}: Enhanced Web and Efficient Knowledge Graph Retrieval for Citation-based Question Answering Systems},
  author={Dehghan, Mohammad and Alomrani, Mohammad and Bagga, Sunyam and Alfonso-Hermelo, David and Bibi, Khalil and Ghaddar, Abbas and Zhang, Yingxue and Li, Xiaoguang and Hao, Jianye and Liu, Qun and Lin, Jimmy and Chen, Boxing and Parthasarathi, Prasanna and Biparva, Mahdi and Rezagholizadeh, Mehdi},
  booktitle={Proceedings of the 62nd Annual Meeting of the Association for Computational Linguistics (Volume 1: Long Papers)},
  pages={14169--14187},
  year={2024},
  doi={10.18653/v1/2024.acl-long.764},
  url={https://aclanthology.org/2024.acl-long.764/}
}

@inproceedings{zhu2020deep,
  author={Zhu, Yanqiao and Xu, Yichen and Yu, Feng and Liu, Qiang and Wu, Shu and Wang, Liang},
  title={{Deep Graph Contrastive Representation Learning}},
  booktitle={ICML Workshop on Graph Representation Learning and Beyond},
  year={2020},
  url={https://arxiv.org/abs/2006.04131}
}

@inproceedings{morris2020tudataset,
  title={TUDataset: A Collection of Benchmark Datasets for Learning with Graphs},
  author={Morris, Christopher and Kriege, Nils M. and Bause, Franka and Kersting, Kristian and Mutzel, Petra and Neumann, Marion},
  booktitle={International Conference on Machine Learning Workshops},
  year={2020},
  url={https://arxiv.org/abs/2007.08663}
}

@inproceedings{yang2016revisiting,
  title={Revisiting Semi-Supervised Learning with Graph Embeddings},
  author={Yang, Zhilin and Cohen, William W. and Salakhutdinov, Ruslan},
  booktitle={International Conference on Machine Learning},
  year={2016},
  url={https://arxiv.org/abs/1603.08861}
}

@inproceedings{kipf2017semi,
  title={Semi-Supervised Classification with Graph Convolutional Networks},
  author={Kipf, Thomas N. and Welling, Max},
  booktitle={International Conference on Learning Representations},
  year={2017},
  url={https://arxiv.org/abs/1609.02907}
}

@inproceedings{velickovic2018graph,
  title={Graph Attention Networks},
  author={Veli\v{c}kovi\'c, Petar and Cucurull, Guillem and Casanova, Arantxa and Romero, Adriana and Li\`o, Pietro and Bengio, Yoshua},
  booktitle={International Conference on Learning Representations},
  year={2018},
  url={https://arxiv.org/abs/1710.10903}
}

@inproceedings{xu2019powerful,
  title={How Powerful are Graph Neural Networks?},
  author={Xu, Keyulu and Hu, Weihua and Leskovec, Jure and Jegelka, Stefanie},
  booktitle={International Conference on Learning Representations},
  year={2019},
  url={https://arxiv.org/abs/1810.00826}
}

@inproceedings{hamilton2017inductive,
  title={Inductive Representation Learning on Large Graphs},
  author={Hamilton, William L. and Ying, Rex and Leskovec, Jure},
  booktitle={Advances in Neural Information Processing Systems},
  year={2017},
  url={https://arxiv.org/abs/1706.02216}
}

@inproceedings{hou2022graphmae,
  title={GraphMAE: Self-Supervised Masked Graph Autoencoders},
  author={Hou, Zhenyu and Liu, Xiao and Cen, Yukuo and Dong, Yuxiao and Yang, Hongxia and Wang, Chunjie and Tang, Jie},
  booktitle={Proceedings of the 28th ACM SIGKDD Conference on Knowledge Discovery and Data Mining},
  year={2022},
  url={https://arxiv.org/abs/2205.10803}
}

@inproceedings{dai2018adversarial,
  title={Adversarial Attack on Graph Structured Data},
  author={Dai, Hanjun and Li, Hui and Tian, Tian and Huang, Xin and Wang, Lin and Zhu, Jun and Song, Le},
  booktitle={Proceedings of the 35th International Conference on Machine Learning},
  series={Proceedings of Machine Learning Research},
  volume={80},
  pages={1115--1124},
  year={2018},
  publisher={PMLR},
  url={https://proceedings.mlr.press/v80/dai18b.html}
}

@inproceedings{zugner2018adversarial,
  title={Adversarial Attacks on Neural Networks for Graph Data},
  author={Z{"u}gner, Daniel and Akbarnejad, Amir and G{"u}nnemann, Stephan},
  booktitle={Proceedings of the 24th ACM SIGKDD International Conference on Knowledge Discovery and Data Mining},
  pages={2847--2856},
  year={2018},
  publisher={ACM},
  doi={10.1145/3219819.3220078}
}

@inproceedings{zugner2019adversarial,
  title={Adversarial Attacks on Graph Neural Networks via Meta Learning},
  author={Z{"u}gner, Daniel and G{"u}nnemann, Stephan},
  booktitle={International Conference on Learning Representations},
  year={2019},
  url={https://openreview.net/forum?id=Bylnx209YX}
}

@inproceedings{bojchevski2019adversarial,
  title={Adversarial Attacks on Node Embeddings via Graph Poisoning},
  author={Bojchevski, Aleksandar and G{"u}nnemann, Stephan},
  booktitle={Proceedings of the 36th International Conference on Machine Learning},
  series={Proceedings of Machine Learning Research},
  volume={97},
  pages={695--704},
  year={2019},
  publisher={PMLR},
  url={https://proceedings.mlr.press/v97/bojchevski19a.html}
}

@inproceedings{xi2021graphbackdoor,
  title={Graph Backdoor},
  author={Xi, Zhaohan and Pang, Ren and Ji, Shouling and Wang, Ting},
  booktitle={30th USENIX Security Symposium},
  pages={1523--1540},
  year={2021},
  publisher={USENIX Association},
  url={https://www.usenix.org/conference/usenixsecurity21/presentation/xi}
}

@inproceedings{dai2023unnoticeable,
  title={Unnoticeable Backdoor Attacks on Graph Neural Networks},
  author={Dai, Enyan and Lin, Minhua and Zhang, Xiang and Wang, Suhang},
  booktitle={Proceedings of the ACM Web Conference 2023},
  pages={2263--2273},
  year={2023},
  publisher={ACM},
  doi={10.1145/3543507.3583392}
}
